\documentclass[alpha-refs]{nbdt-article}

\fancypagestyle{newstyle}{
\fancyhf{} 
\fancyfoot[R]{\vspace{0.1in} \small \thepage}

}

\usepackage{siunitx}
\usepackage{amsmath}
\usepackage{url}
\usepackage[american]{babel}  
\usepackage{csquotes}

\papertype{Original Article}

\title{A study of animal action segmentation algorithms across supervised, unsupervised, and semi-supervised learning paradigms}
\date{}

\newcommand{\bb}[1]{\mathbf{#1}}
\newcommand{\Eold}[1]{\mathbb{E}#1}
\newcommand{\E}[2]{\mathbb{E}_{#1}\left[#2\right]}

\newcommand{\KL}[1]{\text{KL}#1}

\newcommand{\poz}{p\left(\bb{z}_1\right)}
\newcommand{\poy}{p\left(y_1\right)}
\newcommand{\ptxgz}{p\left(\bb{x}_t | \bb{z}_t\right)}
\newcommand{\ptzgzmoy}{p\left(\bb{z}_t | \bb{z}_{t-1}, y_t\right)}

\newcommand{\ptygymozmo}{p\left(y_t | y_{t-1}, \bb{z}_{t-1}\right)}

\newcommand{\qtygxs}{q\left(y_t | \bb{x}_{\mathcal{T}_t}\right)}
\newcommand{\qtzgxsy}{q\left(\bb{z}_t | \bb{x}_{\mathcal{T}_t}, y_t\right)}
\newcommand{\qtzgxsyt}{q\left(\bb{z}_t | \bb{x}_{\mathcal{T}_t}, \tilde{y}_t\right)}
\newcommand{\qtmozgxsy}{q\left(\bb{z}_{t-1} | \bb{x}_{\mathcal{T}_{t-1}}, y_{t-1}\right)}
\newcommand{\qtmoygxs}{q\left(y_{t-1} | \bb{x}_{\mathcal{T}_{t-1}}\right)}

\newcommand{\normal}[3]{\mathcal{N}\left(#1 | #2, #3\right)}

\newcommand{\datasetl}{\mathcal{D}_l}
\newcommand{\datasetu}{\mathcal{D}_u}

\newcommand{\scubed}{S$^3$LDS}

\author[1]{Ari Blau}
\author[2]{Evan S. Schaffer}
\author[3]{Neeli Mishra}
\author[4]{Nathaniel J. Miska}
\author[5\authfn{1}]{International Brain Laboratory}
\author[1,3]{Liam Paninski}
\author[3]{Matthew R. Whiteway}

\contrib[\authfn{1}]{A list of consortium members appears in Appendix B.}

\affil[1]{Department of Statistics, Columbia University}
\affil[2]{Icahn School of Medicine, Mount Sinai}
\affil[3]{Zuckerman Institute, Columbia University}
\affil[4]{Sainsbury Wellcome Centre, University College London}
\affil[5]{International consortium}

\corraddress{Ari Blau, Department of Statistics, Columbia University}
\corremail{ari.blau@columbia.edu}

\begin{document}

\maketitle
\begin{abstract}
\normalsize
Action segmentation of behavioral videos is the process of labeling each frame as belonging to one or more discrete classes, and is a crucial component of many studies that investigate animal behavior. A wide range of algorithms exist to automatically parse discrete animal behavior, encompassing supervised, unsupervised, and semi-supervised learning paradigms. These algorithms –- which include tree-based models, deep neural networks, and graphical models –- differ widely in their structure and assumptions on the data. Using four datasets spanning multiple species –- fly, mouse, and human –- we systematically study how the outputs of these various algorithms align with manually annotated behaviors of interest. Along the way, we introduce a semi-supervised action segmentation model that bridges the gap between supervised deep neural networks and unsupervised graphical models.
We find that fully supervised temporal convolutional networks with the addition of temporal information in the observations perform the best on our supervised metrics across all datasets.

\end{abstract}


\section{Introduction}

\begin{figure}[t!]
\centering
\includegraphics[width=1\linewidth]{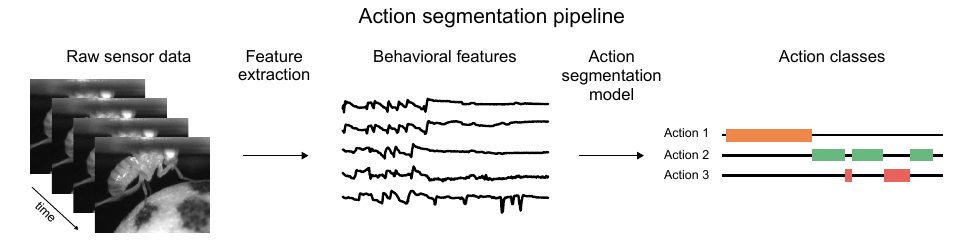}
\vspace{-0.05in}
\caption{\textbf{Overview of the action segmentation pipeline.} 
Raw sensor data (e.g. video) is collected, then features are extracted (e.g. pose estimates), then an action segmentation model is trained to map those features to a discrete behavioral class for each frame.
}
\vspace{-0.05in}
 \label{fig:intro}
\end{figure}

Action segmentation of video data is a process that classifies discrete animal behaviors given a set of spatiotemporal video features. It is an indispensable tool for quantifying natural animal behavior across a range of experimental paradigms, from spontaneous behaviors in open arenas to complex social interactions~\citep{anderson2014toward, pereira2020quantifying, von2021big}. This procedure begins with the collection of raw behavioral data during an experiment, typically with video cameras or motion-capture sensors. The next step is to reduce the dimensionality of the video data using pose estimation~\citep{mathis2018deeplabcut, pereira2019fast, biderman2023lightning}, autoencoders~\citep{batty2019behavenet,whiteway2021partitioning}, or other techniques~\citep{Bohnslav2020deepetho, sun2024video}. Finally, an action segmentation model finds discrete behaviors from the low-dimensional representation (Fig.~\ref{fig:intro}).

\begin{figure}[t!]
\centering
\includegraphics[width=1\linewidth]{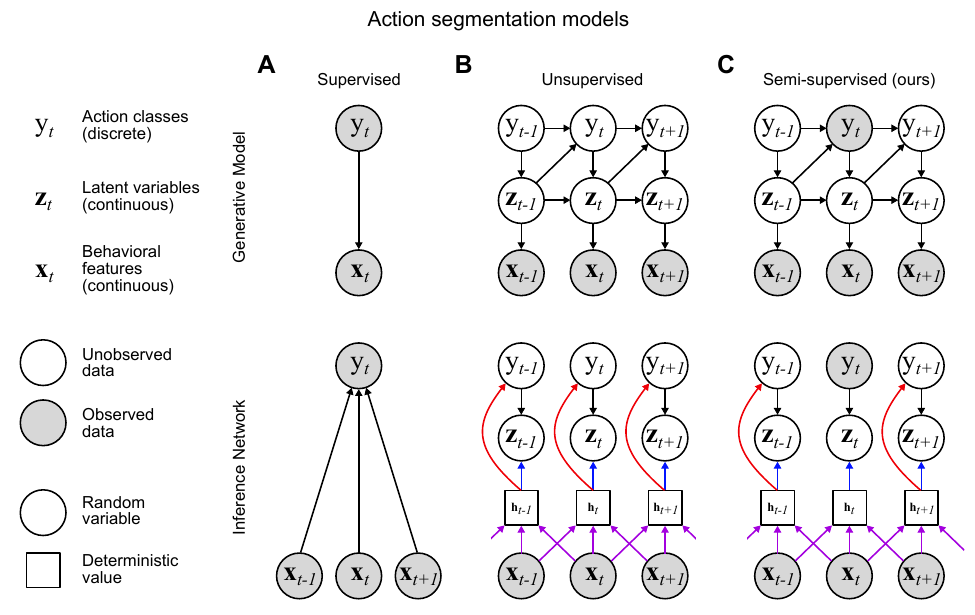}
\vspace{-0.05in}
\caption{\textbf{Overview of action segmentation models.} 
\textbf{A}: \textit{Top}: Graphical model for supervised classification. Both discrete states $y_t$ and poses $\bb{x}_t$ are observed. \textit{Bottom}: Inference network for the supervised model. We use a window of observed behavioral features for state prediction.  
\textbf{B}: \textit{Top}: Graphical model for an unsupervised recurrent switching dynamical system. The set of discrete states $\{y_t\}$ and continuous latents $\{\bb{z}_t\}$ are unobserved. \textit{Bottom}: The inference network uses a window of observed behavioral features to create a deterministic hidden representation $\bb{h}_t$ (purple arrows); this is then used to predict the continuous latents $\bb{z}_t$ (blue arrows) and discrete latents $y_t$ (red arrows). Note that the purple and red arrows together define a classifier for the discrete state at each time step.
\textbf{C}: Graphical model and inference network for a semi-supervised recurrent switching dynamical system. A subset of the discrete states are observed. During inference, the observed discrete state is used for the inference of $\bb{z}_t$ when possible.
}
\vspace{-0.05in}
 \label{fig:models}
\end{figure}

Supervised action segmentation (Fig.~\ref{fig:models}A) requires the experimenter to annotate a subset of frames that contain behaviors of interest, such as grooming or sniffing. Then a classifier is trained to match each frame (or its low-dimensional representation) with the corresponding label~\citep{jhuang2010automated, kabra2013jaaba, kramida2016automated, murari2019recurrent, nguyen2019applying, ravbar2019automatic,  van2020deep, Sturman2020, Bohnslav2020deepetho, segalin2021mouse, gabriel2022behaviordepot, goodwin2024simple}. As the scale of behavioral data continues to grow~\citep{gomez2014big, von2021big}, it becomes infeasible to densely label behaviors in every video. Therefore, it is crucial to develop action segmentation models that perform well with sparsely labeled data. Furthermore, we would like to develop techniques that also take advantage of the vast amounts of unlabeled data~\citep{sun2020task, azabou2024relax}. 

Unsupervised action segmentation models (Fig.~\ref{fig:models}B) are a complementary approach that do not require any hand labels \citep{berman2014mapping, wiltschko2015mapping, johnson2016composing, hsu2021bsoid, luxem2022identifying, weinreb2023keypoint} and perform clustering on the low-dimensional behavioral representation \citep{datta2019computational}. These unsupervised models are more scalable than their supervised counterparts since they do not require annotations for training. Another benefit of this approach is its ability to discover new behaviors that are not pre-defined by the experimenter \citep{datta2019computational, pereira2020quantifying}. However, there may be certain behaviors of particular interest to downstream users, and unsupervised models cannot guarantee that they cluster these behaviors accurately.

The recent proliferation of approaches to animal action segmentation raises obvious practical questions: how do these approaches compare, and what are their trade-offs? To aid in this endeavor, we introduce an action segmentation model that bridges the gap between supervised deep neural networks and unsupervised graphical models through the use of semi-supervised state space models (Fig.~\ref{fig:models}C). Graphical models, particularly switching linear dynamical systems, are an increasingly ubiquitous approach to unsupervised animal action segmentation \citep{wiltschko2015mapping, johnson2016composing, buchanan2017quantifying, costacurta2022distinguishing, weinreb2023keypoint}. These models posit that each discrete action is defined by a set of linear dynamics, and infer which set of dynamics best describes the observations at each time point \citep{datta2019computational}. We utilize such a model and bias it towards interpretable solutions by providing a small set of hand labels during training {(see \cite{ey2016learn} for a similar RNN-based approach)}. Our inference procedure leads to a deep neural network classifier that allows us to make direct comparisons with fully supervised classifiers. We refer to the resulting model and inference technique as a semi-supervised switching linear dynamical system (\scubed).

Another important consideration for the animal action segmentation problem is the behavioral representation used as input to a given model. Much previous work with supervised models uses a host of quantities derived from pose estimates, like distances and angles between different keypoints \citep{Sturman2020, segalin2021mouse, goodwin2024simple}. Much of the unsupervised literature simply uses the pose estimates themselves (perhaps converting to ego-centric coordinates first) and allow the model to perform the necessary transformations \citep{luxem2022identifying, weinreb2023keypoint} (though see \cite{berman2014mapping, hsu2021bsoid} for approaches that involve feature engineering). We would also like to know how simple choices about the behavioral representation -- such as whether or not temporal information contained in velocity or acceleration measurements is included -- impacts the performance of these models. 

To address these questions we use an array of models, including \scubed, to evaluate four behavioral datasets: a head-fixed fly spontaneously behaving on an air-supported ball \citep{schaffer2023spatial}, a mouse freely moving around an open arena \citep{Sturman2020}, a head-fixed mouse performing a perceptual decision-making task from the International Brain Laboratory \citep{international2023brain}, and a human gait dataset (HuGaDB) \citep{chereshnev2017hugadb}.
For the first three datasets, we experiment with both position features (static pose information) and position-velocity features (including the velocity of each keypoint). The HuGaDB data is comprised of three dimensional accelerometer and gyroscope data collected from inertial measurement units (IMUs).

It is challenging to comprehensively compare the performance of such a wide range of models, and we therefore limit ourselves to evaluating how much the outputs of the models align with human-annotated behaviors. This supervised metric will of course miss many of the nuances provided by these models, especially those which are fully unsupervised. We find that the \scubed~offers improvements over supervised models when using the position features as input, but these advantages disappear once we include the velocity features. In the latter case, a fully supervised temporal convolutional network (TCN) performs best across all datasets 
\citep{sun2021multi,schaffer2023spatial,everett2024coordination}. We also compare the \scubed~and fully supervised models to keypoint-MoSeq \citep{weinreb2023keypoint}, a closely related yet fully unsupervised action segmentation method, and find that the models trained with discrete behavior labels learn an improved latent representation, as measured by several supervised cluster quality metrics.

\section{Results}

We first provide a brief overview of the different types of supervised models that we compare, and establish which one we use as our baseline. We explore how performance is affected by the choice of behavioral features used by the various models. Next, we introduce our proposed semi-supervised framework, \scubed, which we compare to our supervised baseline using supervised classification metrics. We also perform several ablation experiments demonstrating which aspects of the semi-supervised model are most important for performance. Finally, we compare the supervised and semi-supervised models to keypoint-MoSeq \citep{weinreb2023keypoint}, an unsupervised model closely related to the \scubed.
{The implementation of the supervised TCN model and all of the semi-supervised models are publicly available in our github repository (\url{https://github.com/ablau100/daart}).}


\subsection{Comparing supervised baselines}

We experiment with three fully supervised models: a temporal convolution network (TCN) \citep{lea2016temporal,lea2017temporal}, random forests \citep{breiman2001random}, and XGBoost \citep{chen2016xgboost}.
TCNs have been competitive on human action segmentation and recognition benchmarks \citep{kim2017interpretable, farha2019ms, filtjens2022skeleton}.
The two tree-based methods are commonly used as supervised classifiers for animal action segmentation \citep{segalin2021mouse, goodwin2024simple}.

Another common question in the literature is how to transform the data before model fitting. Previous work has proposed various hand-engineered features \citep{berman2014mapping, ravbar2019automatic, hsu2021bsoid, segalin2021mouse, goodwin2024simple} and learned features \citep{sun2020task, Bohnslav2020deepetho, azabou2024relax}. While a thorough characterization of this choice is beyond the scope of this study, we test two scenarios: modeling with position features (no temporal information) versus position-velocity features (contains temporal information).

We find the TCN achieves the best performance with both feature types (Fig.~\ref{fig:results_super_baselines}). We also find that temporal information in the features is useful for all methods.
We select the TCN to use as our baseline supervised model, not only because of its superior performance, but also because of its compatibility to function as an inference network in the deep graphical models we introduce next.

\subsection{Introducing a semi-supervised action segmentation model}

Next we move to semi-supervised models, which utilize both labeled and unlabeled data. We focus on the switching linear dynamical system (SLDS) and develop a new amortized variational inference scheme. The SLDS is commonly used to model complex nonlinear behavior in dynamical systems \citep{guy1970state,bing1978state,ham1990time}. 
Several studies have used the SLDS (and the related auto-regressive hidden Markov model, or ARHMM) to model animal behavior \citep{wiltschko2015mapping, johnson2016composing, buchanan2017quantifying, markowitz2018striatum, batty2019behavenet, wilt2020pharm, whiteway2021partitioning, costacurta2022distinguishing, lee2023switching, markowitz2023spontaneous, weinreb2023keypoint}.
The motivation to construct our semi-supervised model from the SLDS is two-fold:
(1) the SLDS is typically utilized as an unsupervised model, and hence allows us to make easy connections between these two learning paradigms; (2) the amortized variational inference scheme will result in a classifier that can be directly compared to the supervised TCN model above.

The standard setup for the generative model of an SLDS is as follows (Fig.~\ref{fig:models}B): let $y_t \in \{0, \ldots, K-1\}$ be the discrete behavioral state at time $t$ with Markovian dynamics, i.e. the state at time $t$ is only dependent on the state at time $t-1$. 

The observed data $\bb{x}_t$ are modeled using piecewise-linear dynamics in a latent space denoted by $\bb{z}_t$:
\begin{eqnarray}
p_{\theta}\left(y_{t} | y_{t-1}\right) &=& \textrm{Cat}\left(y_{t} |\pi_\theta\left({y_{t-1}}\right)\right) \label{eq:py} \\
p_{\theta}\left(\bb{z}_{t} | \bb{z}_{t-1}, {y_{t}} \right) &=& \normal{\bb{z}_{t}}{A_{y_t} \bb{z}_{t-1} + b_{{y}_{t}}}{Q_{y_{t}}} \label{eq:pz_zy} \\ 
p_{\theta}\left(\bb{x}_t | \bb{z}_{t}, y_t\right) &=& \normal{\bb{x}_t}{C_{y_t}{\bb{z}_t} + d_{y_t}}{S_{y_t}}, \label{eq:px_zy}
\end{eqnarray}
where $t \in \{1,2,\dots , T\}$. $A_{y_t}$ and $b_{y_t}$ define the latent linear dynamics associated with state $y_t$; $C_{y_t}$ and $d_{y_t}$ represent the linear mapping from the continuous latents to the observations; and $Q_{y_t}$ and $S_{y_t}$ represent noise covariance matrices of the latent and observed data. 
In other words, at each time point, the model selects one set of the $K$ linear dynamics (which each define a discrete action class; Eq.~\ref{eq:py}) that operate in a continuous latent space, evolves the latents forward in time using those linear dynamics (Eq.~\ref{eq:pz_zy}), then projects those latents to the observations (the behavioral features; Eq.~\ref{eq:px_zy}).

The recurrent SLDS (rSLDS) builds on the SLDS by conditioning $y_{t}$ on $\bb{z}_{t-1}$ in addition to $y_{t-1}$ \citep{linderman2016rslds} (Fig.~\ref{fig:models}B). This recurrence allows the model to more flexibly switch between discrete states based on the trajectory through the latent space, rather than just static switching probabilities. The new generative model therefore replaces Eq.~\ref{eq:py} with
\begin{equation}
  p_\theta\left(y_{t} | y_{t-1}, \bb{z}_{t-1}\right) = \textrm{Cat}\left(y_{t} |\pi_\theta\left(y_{t-1}, \bb{z}_{t-1}\right)\right). \label{eq:py_z}
\end{equation}

In both the SLDS and rSLDS, it is assumed the discrete action classes are all unobserved. We propose to allow the model access to labels for the discrete action classes for a small number of time points (Fig.~\ref{fig:models}C, \textit{top}), resulting in a semi-supervised SLDS (\scubed). We develop an efficient and flexible amortized variational inference strategy that incorporates both labeled and unlabeled data. 
Importantly, the proposed inference strategy leads to the definition of an approximate posterior $q_{\phi}\left(y_t | \bb{x}_{1:T}\right)$ which also serves as a classifier \citep{willetts2020semi}, predicting the discrete action class $y_t$ from the observed data $\bb{x}_{1:T}$ (Fig.~\ref{fig:models}C, \textit{bottom}). We implement this approximate posterior using the TCN model of the previous section, allowing us to directly assess the benefits of training this classifier network with both labeled and unlabeled data. {This approach also leads to training and inference speeds that are substantially faster than the non-amortized approach employed by keypoint-MoSeq ($\sim$4x and $\sim$2x speedups, respectively; Fig.~\ref{fig:time_chart}).} For more details, see the Methods section.

\subsection{Semi-supervised and supervised model comparisons}

We start by comparing supervised and semi-supervised models using a supervised classification metric, and report several ablation studies along the way that highlight the role of various components of the \scubed. Due to our choice of the TCN for the \scubed~inference network, we use a TCN with the exact same architecture as our supervised baseline. 

\begin{figure}[t!]
\centering
\includegraphics[width=\linewidth]{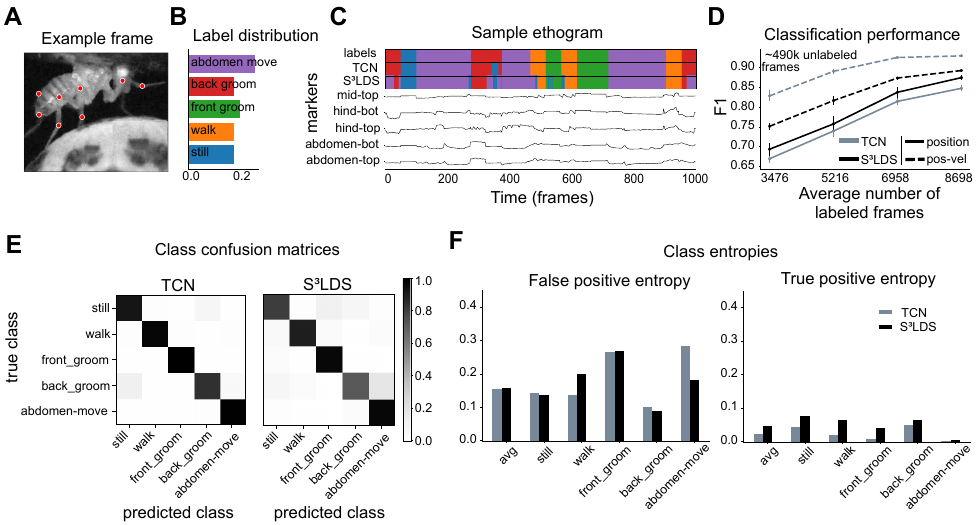}
\vspace{-0.05in}
\caption{\textbf{Supervised vs semi-supervised results for the head-fixed fly.}
\textbf{A}: Example frame of the fly, overlaid with pose markers.
\textbf{B}: Proportion of each labeled behavior in the training dataset.
\textbf{C}: Sample of ground truth labels, along with predictions from both the TCN and the \scubed~models. Below is a subset of the corresponding features used as inputs to the models.
\textbf{D}: F1 scores for the TCN and \scubed~models. We show results for the position features (solid lines) as well as the position-velocity features (dashed lines). Adding velocity improves performance for both models. {The number of unlabeled frames used in the models with the smallest number of labeled frames is displayed in the upper right corner of the graph; this number decreases as we add labels for each consecutive set of models}. Error bars represent the standard deviation of the F1 scores over five subsamples of the training data.
\textbf{E}: Confusion matrices for the TCN and \scubed~models. 
\textbf{F}: Average entropy of the false positives (left) and true positives (right) for both models. Entropy results for the other datasets are shown in Fig.~\ref{fig:results_super_others_detail}.
Panels E and F show results from the models trained on all labeled frames with position-velocity features.
}
\vspace{-0.05in}
\label{fig:results_super_fly}
\end{figure}

We begin with the head-fixed fly dataset (Fig.~\ref{fig:results_super_fly}A), which contains five annotated behaviors: still, walk, front groom, back groom, and abdomen move (Fig.~\ref{fig:results_super_fly}B,C). We extract pose estimates from the videos using Lightning Pose and the Ensemble Kalman Smoother post-processor \citep{biderman2023lightning}, and use these position features as input to the TCN and S$^3$LDS. We train models with an increasing number of labeled videos (five networks trained per condition, each using a random subset of labeled videos). The \scubed~uses all available unlabeled frames. We find that performance, as measured by F1 (the harmonic mean of accuracy and precision; higher is better, with a maximum value of 1.0) improves with increasing labels. The \scubed~outperforms the TCN across all labeled data amounts (Fig.~\ref{fig:results_super_fly}D, solid lines) using these position features, demonstrating how the addition of unlabeled frames can improve supervised classification performance.

While modeling the raw pose estimates is intuitive, the SLDS framework can accommodate any type of behavioral features, and has been used, for example, with depth cameras \citep{wiltschko2015mapping, johnson2016composing, markowitz2018striatum, wilt2020pharm, costacurta2022distinguishing, markowitz2023spontaneous}, latent representations from autoencoders \citep{batty2019behavenet, whiteway2021partitioning}, and other behavioral features \citep{buchanan2017quantifying}. While our goal here is not to determine the optimal behavioral representation for our models, we find that simply concatenating the pose vector $\bb{x}_t$ with its velocity $\bb{x}_t - \bb{x}_{t-1}$ leads to a new behavioral representation $\tilde{\bb{x}}_t = \left[\bb{x}_t; \bb{x}_t-\bb{x}_{t-1}\right]$ of position-velocity features which outperforms the position features $\bb{x}_t$ in both model types (Fig.~\ref{fig:results_super_fly}D, dashed lines). With this new behavioral representation, the performance improvements due to unlabeled frames disappear.

\begin{figure}[t!]
\centering
\includegraphics[width=\linewidth]{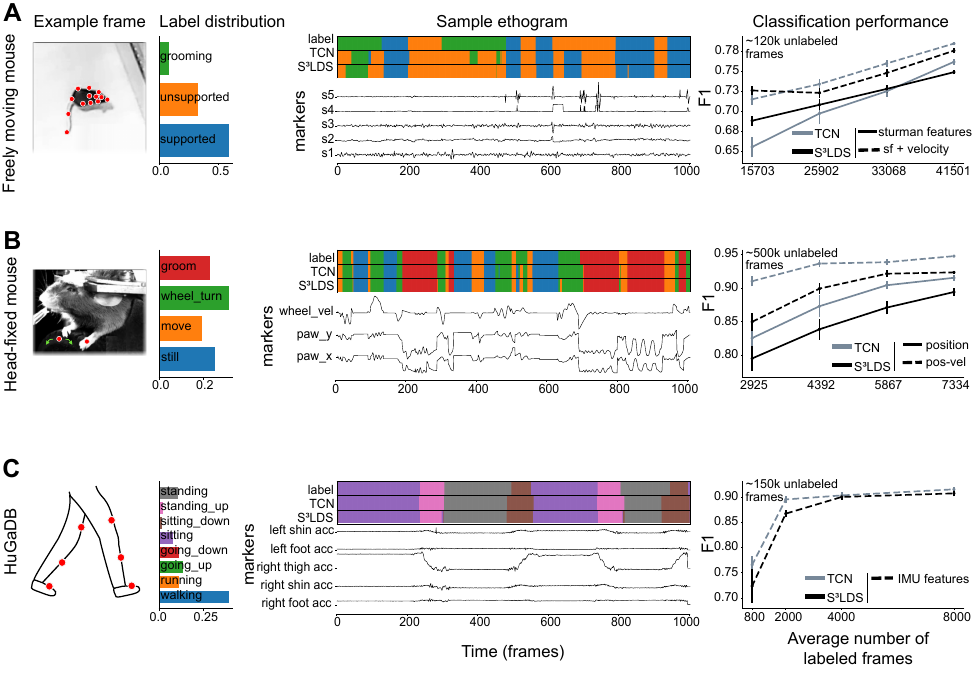}
\vspace{-0.05in}
\caption{\textbf{Supervised vs semi-supervised results across datasets.}
Conventions as in Fig.~\ref{fig:results_super_fly}. As in the head-fixed fly, we find that using position-velocity features improves performance over the position features across both model types, and in all datasets the TCN performs best.
\textbf{A}: Results on the freely moving mouse dataset. Rather than using the raw poses, we compute the features introduced in \cite{Sturman2020}. These features compute transformations on the poses, including distances and angles between different groups of keypoints.
\textbf{B}: Results on the head-fixed mouse dataset.
\textbf{C}: Results on the HuGaDB dataset. The data is collected from sensors that already contain velocity data, so we only use one set of features.
}
\vspace{-0.05in}
\label{fig:results_super_others}
\end{figure}

To further investigate this, we next look at a snippet of model predictions using $\tilde{\bb{x}}_t$ (Fig.~\ref{fig:results_super_fly}C). One observation is that \scubed~makes mistakes across a wider range of behavior classes. We compute the confusion matrices for each model across all test videos, but do not see any obvious structural differences between the models (Fig.~\ref{fig:results_super_fly}E). However, another observation is that the \scubed~predictions contain some rapid state switches. One cause of these rapid switches would be two or more classes whose probabilities are similar, and small amounts of noise can lead to switches between these states. To test this hypothesis, we computed the entropy of the discrete state probabilities for each action class. The models are evenly matched for the false positives (Fig.~\ref{fig:results_super_fly}F). However, for true positives the entropy for the \scubed~is higher across all classes, indicating the semi-supervised model is not as confident as the supervised one.
{We find similar patterns in the remaining datasets: unlabeled data can help when using position features, and hurts when using position-velocity features (Figs.~\ref{fig:results_super_others},~\ref{fig:results_super_others_detail}).}

{To further analyze how the inclusion of unlabeled frames affects performance, we perform an ablation study keeping the number of labeled frames constant while increasing the number of unlabeled frames (Figs.~\ref{fig:unlabeled_m},~\ref{fig:unlabeled_pv}).
We find that adding more unlabeled data to our semi-supervised model improves performance by a small amount when using the position features, but additional unlabeled frames actually decrease performance when using position-velocity features. These results are in agreement with the above conclusion that unlabeled frames can improve upon the supervised TCN, but only when using position features (e.g. Fig.~\ref{fig:results_super_fly}D).}

The finding that inclusion of temporal information in the observations leads to improved classification performance raises a related question: are the temporal components of the \scubed~necessary for accurate action segmentation? To answer this question we perform an ablation study. We start with the semi-supervised Gaussian Mixture Deep Generative Model (GMDGM) \citep{willetts2020semi}, a static analogue of the S$^3$LDS. 
The GMDGM was developed to model static data such as images, and as such neither the approximate posteriors nor the generative model take temporal information into account. As in the S$^3$LDS, we allow for some of the discrete states to be observed (Fig.~\ref{fig:models_ablations}A). To bridge the gap between the GMDGM and the \scubed~we introduce the GMDGM-TCN, which uses the static generative model of the GMDGM and the temporal context-aware approximate posterior of the S$^3$LDS (Fig.~\ref{fig:models_ablations}B). The GMDGM and GMDGM-TCN thus highlight the effects of ablating different aspects of the dynamic structure in the S$^3$LDS.

Across all datasets, we find the GMDGM performs much worse than the temporal models when the observations do not contain temporal information (Fig.~\ref{fig:results_super_ablations}).
Except for the fly dataset, the GMDGM performance is improved by adding temporal information into the inference network (the GMDGM-TCN). Interestingly, adding temporal information to the generative model (the S$^3$LDS) does not meaningfully improve performance across any of the datasets, regardless of the behavioral features used. We conclude that incorporating temporal information in the observations and the approximate posterior provide complementary improvements in model performance, and incorporating temporal information into the generative model does not provide additional gains.



Finally, we look at the role of nonlinearities in the S$^3$LDS. In our implementation we use linear transformations for the recurrent state transitions (Eq.~\ref{eq:py_z}) and the  continuous latent space dynamics (Eq.~\ref{eq:pz_zy}). We also implement a semi-supervised \textit{non}linear dynamical system (S$^3$NLDS) where we replace each of these linear transformations with a one-hidden-layer dense neural network (Methods). The flexibility of our amortized variational inference scheme allows for such changes to the generative model with no corresponding changes in the inference. We find that -- at least for these datasets -- there is no benefit to using nonlinearities, regardless of the behavioral representation (Fig.~\ref{fig:results_super_nonlinearities}).

\subsection{Adding sparse labels to an unsupervised method modifies the latent space}

We now turn to comparisons between the \scubed~and a closely related unsupervised model, keypoint-MoSeq \citep{weinreb2023keypoint}. Keypoint-Moseq is also in the class of SLDS models, but differs in several key aspects. Importantly, keypoint-MoSeq uses a different prior on the discrete states that allows the model to choose the 
number of states from the data; this also requires a very different inference approach than the one developed for the \scubed. Despite these differences (among others), we are interested in comparing the discrete states uncovered by these models. 

\begin{figure}[t!]
\centering
\includegraphics[width=\linewidth]{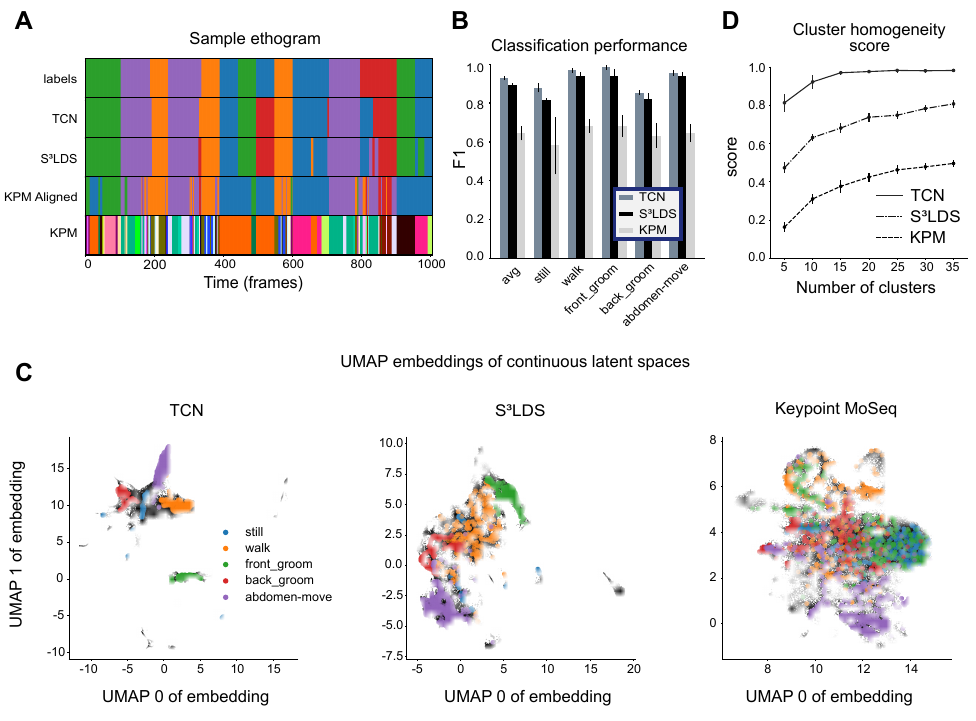}
\vspace{-0.05in}
\caption{\textbf{Supervised and semi-supervised latent spaces more closely align with labels than unsupervised latents (head-fixed fly).}
All models use position-velocity features and all available training videos for the head-fixed fly dataset.
\textbf{A}: The top row shows a segment of ground truth labels. The following two rows show predictions from the TCN and \scubed~models. The third row shows the state outputs of keypoint-MoSeq (KPM), aligned to the ground truth class with highest overlap on the training data. The final row shows the raw state outputs of keypoint-MoSeq.
\textbf{B}: F1 scores for the TCN, \scubed~and KPM models. Error bars represent the standard deviation of the F1 scores over five trained models (different initialization seeds).
\textbf{C}: 2D UMAP embedding of continuous latents colored by discrete labels for three different models. 
\textbf{D}: The addition of hand labels produces more homogeneous clusters in the models’ latent spaces. Error bars represent the standard deviation of the cluster scores over five models. We use a range of cluster numbers to show that cluster scores are not biased by cluster size.
}
\vspace{-0.05in}
\label{fig:results_kpm_fly}
\end{figure}

We fit keypoint-Moseq on the position-velocity fly features, and find 37 discrete states. We then match each of these states to the labeled action class that contains the most overlap in the training data, and find good overlap with the ground truth labels on held-out test data (Fig.~\ref{fig:results_kpm_fly}A). We quantify this overlap with the same F1 score as before, and find that keypoint-MoSeq achieves competitive results, but unsurprisingly the TCN and \scubed~-- which both have access to the hand labels during training -- perform better on this supervised metric (Fig.~\ref{fig:results_kpm_fly}B).

\begin{figure}[t!]
\centering
\includegraphics[width=\linewidth]{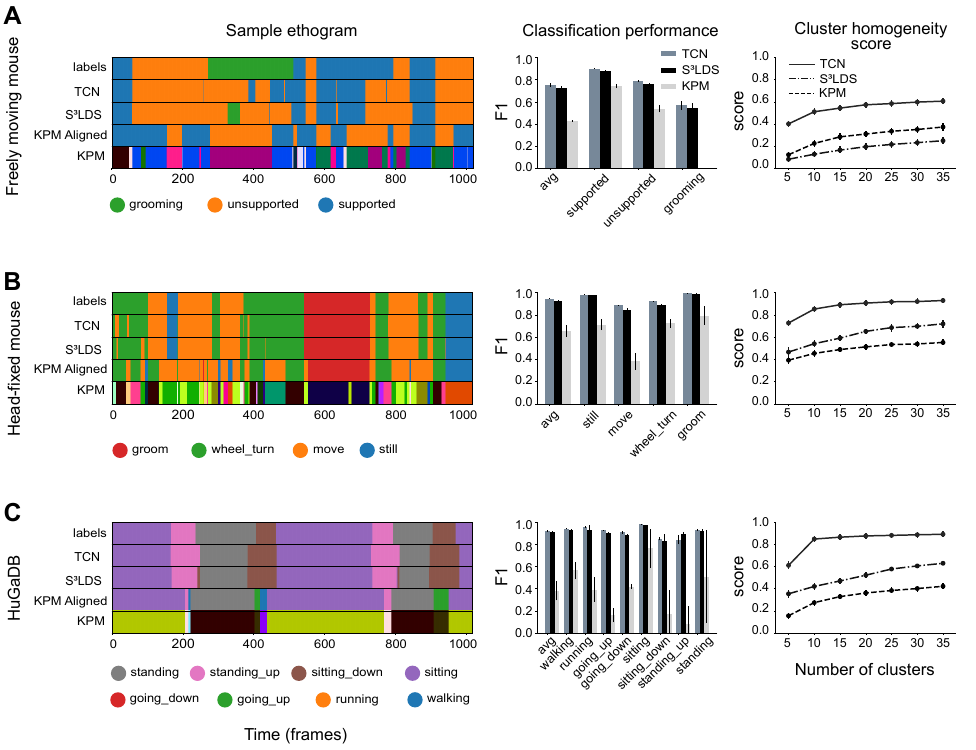}
\vspace{-0.05in}
\caption{\textbf{Keypoint-MoSeq performance on non-fly datasets: position-velocity features.}
Models are trained with position-velocity features for all datasets. The mouse datasets (panels A and B) use position-velocity features, while the HuGaDB dataset uses inertial sensor data (panel C). Other conventions as in Fig.~\ref{fig:results_kpm_fly}. As in the fly dataset, we find the TCN, which is purely supervised, achieves the highest alignment of the latent space with the ground truth labels as measured by the cluster homogeneity score.
}
\vspace{-0.05in}
 \label{fig:results_kpm_others}
\end{figure}

It is also possible to investigate the effects of hand labels on the continuous latent space of each model type. The latent embeddings are computed by passing the observations through the trained encoders for the TCN and \scubed, and performing inference of the continuous latents in keypoint-MoSeq. We first project the latents of each model into a 2D space using UMAP \citep{mcinnes2018umap} (Fig.~\ref{fig:results_kpm_fly}C). The points with corresponding hand labels are colored, revealing that similar behaviors are clustered together across all model types. To quantify this observation, we next perform k-means clustering in the original latent space for each model. We then compute a cluster homogeneity score that measures the extent to which the k-means clusters contain data points from a single behavior class~\citep{rosenberg2007v}. The S$^3$LDS achieves a higher score than keypoint-MoSeq, demonstrating that the presence of hand labels during training does in fact shift the continuous latent space to be more aligned with the labeled behavior classes (Fig.~\ref{fig:results_kpm_fly}D). The TCN, which is purely supervised, scores much higher than the other two models since its only objective is to properly separate the hand labels. These observations are replicated across the other datasets (Fig.~\ref{fig:results_kpm_others}) and with various other cluster quality metrics (Fig.~\ref{fig:results_kpm_other_metrics_pv}). We show results on the non-temporal features in the supplemental material (Figs.~\ref{fig:results_kpm_other_m},~\ref{fig:results_kpm_other_m_f1},~\ref{fig:results_kpm_other_metrics_m}).

We close by noting the F1 score and cluster quality metrics are all based on hand labels; a purely unsupervised method like keypoint-MoSeq is not trained to optimize these types of metrics. Rather, keypoint-MoSeq (and related methods) finds action classes defined by statistical regularities in the behavioral dynamics. Our finding here is that these statistical regularities may not match human-defined behavioral classes as precisely as models trained specifically for this task.

\section{Discussion}
We surveyed a range of animal action segmentation models spanning supervised, semi-supervised, and unsupervised learning paradigms. 
Across four behavioral datasets, we established that supervised TCN models trained with both static and dynamic pose information performed the best on supervised classification metrics.
We also proposed the \scubed, a semi-supervised method for animal action segmentation that incorporates a small amount of discrete labels. With the \scubed~we were able to selectively ablate different aspects of the model and assess how they impacted performance; we found that including temporal information in the inference network and the observations is critical, but temporal information in the generative model did not lead to performance improvements.
Furthermore, we compared these models to the unsupervised keypoint-MoSeq, and showed how the inclusion of labeled data during training shapes the latent spaces of these models to be more aligned with labeled classes.

A straightforward but important conclusion from our work is that the behavioral representation used for action segmentation -- whether supervised, unsupervised, or something in between -- plays a crucial role in the performance of the algorithm. \cite{Sturman2020} did not use raw pose estimation outputs for action segmentation, but rather computed a set of derived quantities like distances and angles between different keypoints. We confirmed that this representation is much more performant than the raw keypoints themselves (data not shown), but also found that additionally using the velocities of these features led to further improvements (even though the TCN-based models explicitly use a temporal window of features as input). But what are the ``best'' set of features for any given dataset? The problem of featurization becomes even more complex in the case of social interactions; for example, both the MARS \citep{segalin2021mouse} and SimBA \citep{goodwin2024simple} packages compute hundreds of behavioral features within and between animals as input to their classifiers. There has been some work on automating this process. \cite{sun2020task} {and \cite{azabou2024relax}} use large amounts of unlabeled data to build an effective feature extractor from pose data, and then use simple, fully supervised classifiers for the final stage. Similarly, \cite{Bohnslav2020deepetho} {and \cite{sun2024video}} use a set of unsupervised feature extractors to bypass pose estimation and directly compute abstract behavioral representations from video frames, which are then fed into fully supervised classifiers.

An important direction for future work is to leverage the \scubed~framework for semi-\textit{un}supervised learning \citep{willetts2020semi, davidson2021semi}; that is, where we provide a small number of labels for some behavioral categories, but also provide the model with capacity to discover completely novel classes simultaneously. Such a model would allow experimenters to precisely quantify known behaviors of interest while also capturing previously unknown behaviors in the data. {Our \scubed~framework naturally allows for this scenario; all that is required is to expand the number of discrete states and not provide any hand labels for the new states. However, it might also be necessary to encourage the model to use these new discrete states during training. For example, the amortized variational inference scheme for a fully unsupervised nonlinear SLDS model proposed by \cite{dong2020collapsed} requires an additional regularizing term in the ELBO that penalizes the KL divergence between
the state posterior and a uniform prior at each time step; otherwise they found the model would sometimes only utilize a single discrete state.} 

Another avenue for future work involves modeling the social interactions between multiple animals. Recent work has modeled the interactions between multiple brain regions using rSLDS-like models \citep{glaser2020recurrent, karniol2023modeling}, and the flexibility of our approach -- as well as the relative ease with which we can label discrete animal behaviors compared to discrete brain states -- make the \scubed~an interesting starting point for such modeling efforts.

{To support this work we have developed a single codebase that is capable of incorporating labels into unsupervised models, as well as unsupervised losses for supervised models, to enable semi-supervised learning. This codebase was used to fit the TCN, GMDGM, and \scubed~models in this manuscript. We hope that the publicly available code (\url{https://github.com/ablau100/daart}) 
will support further exploration of semi-supervised learning in action segmentation models.}



\paragraph{Acknowledgements} We thank Peter Latham and Matteo Carandini for helpful discussions. This work was supported by the following grants: Gatsby Charitable Foundation GAT3708, Kavli Foundation, NIH U19NS123716, NSF 1707398, The NSF AI Institute for Artificial and Natural Intelligence, Simons Foundation and the Wellcome Trust 216324. The funders had no role in study design, data collection and analysis,
decision to publish, or preparation of the manuscript.

\paragraph{Code availability} \label{sec:code} The code for the TCN, \scubed, and GMDGM is available at \url{https://github.com/ablau100/daart} under the GPL-3.0 license.

\paragraph{Data availability} 
{For all datasets, we make the data available in the format required by our codebase, including videos, pose estimates, features, and discrete state annotations. For all but the head-fixed fly dataset, we also link to the unprocessed, publicly available data.}
\vspace{1em}

{\textbf{Head-fixed fly}: the processed data is available at \url{https://figshare.com/articles/dataset/Semi-supervised_animal_action_segmentation_head-fixed_fly/27478047?file=50195331}.}
\vspace{1em}

\textbf{Freely moving mouse}
\vspace{-2.5em}
\begin{itemize}
    \item The pose estimates and discrete state annotations are available at \url{https://github.com/ETHZ-INS/DLCAnalyzer}; the raw videos are available at \url{https://zenodo.org/records/3608658}
    \item {The processed data is available at \url{https://figshare.com/articles/dataset/Semi-supervised_animal_action_segmentation_freely_moving_mouse/27477885?file=50155506}}
\end{itemize}

\textbf{Head-fixed mouse}
\vspace{-2.5em}
\begin{itemize}
    \item To access the raw videos, pose estimates, and wheel traces for the IBL datasets, see the documentation at
\url{https://int-brain-lab.github.io/ONE/FAQ.html#how-do-i-download-the-datasets-cache-for-a-specific-ibl-paper-release} and use the tag \texttt{2024\_Q2\_Blau\_et\_al}
    \item {The processed data is available at \url{https://figshare.com/articles/dataset/Semi-supervised_animal_action_segmentation_head-fixed_mouse/27479760?file=50155986}}
\end{itemize}

\textbf{Human Gait Database}
\vspace{-2.5em}
\begin{itemize}
    \item The IMU sensor data can be found at \url{https://www.dropbox.com/scl/fi/jsq0mr26rthrzy64vkkjc/HuGaDB-v2.zip?rlkey=101j8lvdktdejm105cf9fpisi&e=1&dl=0}
    \item {The processed data is available at \url{https://figshare.com/articles/dataset/Semi-supervised_animal_action_segmentation_HuGaDB/27479808?file=50156106}}
\end{itemize}
\newpage

\section{Methods}

\subsection{Datasets}

\paragraph{Head-fixed fly} A head-fixed fly engaged in spontaneous behaviors on an air-supported ball \citep{schaffer2023spatial}. A camera captured a side view of the fly at 70 Hz. We tracked eight points across the legs and abdomen using five supervised Lightning Pose networks, followed by Ensemble Kalman Smoothing \citep{biderman2023lightning}.

We consider five behavior categories: still, walk, front groom, back groom, and abdomen move. {A single annotator} labeled behaviors in chunks of 50 to several hundred contiguous time points using the DeepEthogram GUI \citep{Bohnslav2020deepetho}. Flies often engage in behaviors for longer than 50 frames, so the selected chunks did not contain any transitions from one behavior to another.
We use five labeled videos for training, and five for testing. The dataset contains 1.01M frames, only 16.6k of which are labeled ($\sim$1.6\%; Table~\ref{tab:labels-fly}). We train the models on 2, 3, 4, and 5 of the training videos to see how our models perform with different amounts of labeled data. For the models trained with five videos, we run each model five times with different random seeds. For the other models, we subsample different permutations of the training videos, i.e. for two videos, we take five random permutations of two videos from the list of all five training videos.

\paragraph{Freely moving mouse} In this publicly available dataset, a mouse freely moved around an open arena \citep{Sturman2020}. A camera captured a top-down view of the mouse at 25 Hz. Thirteen points were tracked across the tail, body, ears, and nose of the mouse{; the paws were not tracked.} Instead of using the raw pose data for observations, we use a set of 21 features computed from the pose data based on distances, angles, and areas, following \cite{Sturman2020}. This provides a position- and orientation-invariant representation that we did not need to account for in the head-fixed fly dataset. {We also include the distance between the centroid of the mouse to the boundary of the arena.}

We consider {all} three behavior categories {provided in the public dataset}: unsupported rearing, supported rearing (when the mouse uses the wall of the arena), and grooming. In the rest of the frames, there is no specific action or behavior displayed, so we do not train any of the models to identify this ``other'' category. We only consider the three specific behaviors described above in our analyses. We use ten labeled videos for training, and ten for testing. The dataset contains 280k frames, 83k of which are labeled ($\sim$29.5\%; Table~\ref{tab:labels-mouse}). We train our models on 4, 6, 8, and 10 videos. We choose the permutations of these videos similar to the head-fixed fly videos.

\paragraph{Head-fixed mouse} This is a publicly available dataset from the International Brain Laboratory \citep{international2023brain}. A head-fixed mouse performed a perceptual decision-making task, using a wheel to indicate its decision \citep{international2021standardized, international2023brain}. Two cameras – ``left'' (60 Hz) and ``right'' (150 Hz) – capture roughly orthogonal side
views of the mouse’s face and upper trunk during each session \citep{ibl2022video}, though we only use the left view. Multiple points are tracked across the paws and face, though for this work we only consider a single tracked point on the paw closest to the camera. In addition to the paw coordinates, we also use the 1D wheel velocity as input to the models. See example videos from one of the sessions here: \url{https://viz.internationalbrainlab.org/app?dset=bwm&pid=94fcff55-2da2-4366-a2c7-2f58c05b54dc&tid=57&cid=598&qc=0#trialviewer}.

We consider four behavioral categories for this single paw: still, moving without turning the wheel, moving while turning the wheel, and grooming. {A single annotator} labeled behaviors using the DeepEthogram GUI \citep{Bohnslav2020deepetho}. We use five labeled videos for training, and five for testing. The dataset contains 1M frames, 14k of which are labeled ($\sim$1.4\%; Table~\ref{tab:labels-ibl}). We train the models on 2, 3, 4, and 5 of the training videos, as described for the other datasets.

\paragraph{Human Gait Database} The Human Gait Database (HuGaDB) is a publicly available action segmentation dataset that contains lower limb activities such as walking, running, and sitting for a total of 18 subjects \citep{chereshnev2017hugadb}. 
HuGaDB measured movements from six inertial measurement units (IMUs) with a 60 Hz sampling frequency. The IMUs were placed on each individual's right and left feet, shins, and thighs. Each inertial sensor tracks the acceleration data and gyroscope data on each of the three x,y,z axes.

We consider eight behavioral categories: walking, running, going up, going down, sitting, sitting down, standing up, and standing. The original dataset contained four additional behaviors that we excluded from our analysis. Two of those behaviors, ``bicycling'' and ``sitting in car,'' were not present in the publicly available dataset. The IMU data for the classes ``up by elevator'' and ``down by elevator'' did not seem to contain measurable variability. The dataset contains 517k frames, all of which are labeled (Table \ref{tab:labels-hugadb}). We train our model on 100, 250, 500, and 1000 labeled frames from each class. For each number of training frames, we randomly select the specified number of labeled frames, and remove the rest of the labels for training.

\subsection{Semi-supervised linear dynamical system (\scubed)}

\subsubsection{\scubed~model formulation} \label{sec:model}

Our starting point is the switching linear dynamical system (SLDS) defined in Eqs.~\ref{eq:py}-\ref{eq:py_z}. The specific model that we implement further generalizes the above model by allowing the transitions between discrete states to be recurrent, and the observation mapping to be nonlinear, leading to a modified recurrent switching linear dynamical system (rSLDS) model \citep{dong2020collapsed,karniol2023modeling}:


\begin{eqnarray}
p_{\theta}\left(\bb{x}_{1:T}, y_{1:T}, \bb{z}_{1:T}\right) &=& p_{\theta}\left(\bb{x}_1 | \bb{z}_1\right)p_{\theta}\left(\bb{z}_1\right)p\left(y_1\right)
\prod_{t=2}^T p_{\theta}\left(\bb{x}_t | \bb{z}_t\right)p_{\theta}\left(\bb{z}_t | \bb{z}_{t-1}, y_t\right)p_{\theta}\left(y_t | y_{t-1}, \bb{z}_{t-1}\right) \\
p_{\theta}\left(\bb{x}_t | \bb{z}_t\right) &=& \normal{\bb{x}_t}{g_{\theta}\left(\bb{z}_t\right)}{S} \label{eq:p_xt_g_zt}\\
p_{\theta}\left(\bb{z}_{t} | \bb{z}_{t-1}, {y_{t}} \right) &=& \normal{\bb{z}_{t}}{A_{y_t} \bb{z}_{t-1} + b_{{y}_{t}}}{Q_{y_{t}}} \label{eq:p_zt_g_ztm1}\\
p_{\theta}\left(y_t | y_{t-1}, \bb{z}_{t-1}\right) &=& \text{Cat}\left(y_t | \text{Softmax}\left(R_{y_{t-1}}\bb{z}_{t-1} + r_{{y}_{t-1}}\right)\right) \label{eq:p_yt_g_ytm1}\\
p\left(\bb{z}_1\right) &=& \normal{\bb{z}_1}{0}{\mathbb{I}} \\
p\left(y_1\right) &=& \text{Cat}\left(y_1 | \pi\right),
\end{eqnarray}
where $R_{y_t}$ and $r_{y_t}$ define the recurrent transition associated with state $y_t$, and $g_{\theta}$ is a one-hidden-layer dense neural network that is shared across all discrete states. For our investigation of nonlinearities (Fig.~\ref{fig:results_super_nonlinearities}), we replaced each of the linear transformations in Eqs.~\ref{eq:p_zt_g_ztm1} and \ref{eq:p_yt_g_ytm1} with a one-hidden-layer dense neural network.

\subsubsection{\scubed~inference and learning}
Switching models contain discrete latent variables, and therefore we cannot simply use the reparameterization trick commonly employed in variational inference schemes. One option is to use a continuous relaxation of the discrete variables such as the Gumbel Softmax \citep{jang2016categorical}, Concrete \citep{maddison2016concrete}, or REBAR \citep{tucker2017rebar}. However, in practice this approach has not been successful with switching \textit{non}linear models (SNLDS) \citep{dong2020collapsed}. Both \cite{dong2020collapsed, karniol2023modeling} use the exact posterior of the discrete states under the generative model, and perform inference using the forward-backward algorithm. We take a different approach and marginalize out the discrete state such that we obtain a soft mixture of dynamics at each time step, similar to the Kalman VAE \citep{fraccaro2017disentangled}.

We define the approximate posteriors for this model in a way that leads to a classifier $q_{\phi}\left(y_t | \bb{x}_{1:T}\right)$, which is trained on both labeled and unlabeled data \citep{willetts2020semi}. We also incorporate temporal information in our approximate posteriors to more closely match the generative model \citep{krishnan2015deep, krishnan2017structured}. 
We propose to do so with a temporal convolution network \citep{lea2016temporal, lea2017temporal}. To begin, let us assume we have no observed labels $y_t$; we define a joint posterior over $\{y_{1:T}, \bb{z}_{1:T}\}$ that factorizes over time:
\begin{equation}
q_{\phi}\left(y_{1:T}, \bb{z}_{1:T} | \bb{x}_{1:T}\right) = \prod_{t=1}^T q_{\phi}\left(y_t, \bb{z}_t | \bb{x}_{\mathcal{T}_t}\right)
\end{equation}
where $\mathcal{T}_t$ defines a set of time points, for instance $\{1, \ldots, T\}$ if we condition on all observations, or $\{t-\tau, \ldots, t+\tau\}$ if we condition on a window of observations centered at time $t$. We further factorize $q_{\phi}\left(y_t, \bb{z}_t | \bb{x}_{\mathcal{T}_t}\right)$ as
\begin{eqnarray}
q_{\phi}\left(y_t, \bb{z}_t | \bb{x}_{\mathcal{T}_t}\right) &=& q_{\phi}\left(y_t | \bb{x}_{\mathcal{T}_t}\right) q_{\phi}\left(\bb{z}_t | \bb{x}_{\mathcal{T}_t}, y_t\right) \label{eq:qyz}\\
q_{\phi}\left(y_t | \bb{x}_{\mathcal{T}_t}\right) &=& \text{Cat}\left(y_t | \pi_{\phi}\left(\bb{x}_{\mathcal{T}_t}\right)\right) \label{eq:qy}\\
q_{\phi}\left(\bb{z}_t | \bb{x}_{\mathcal{T}_t}, y_t\right) &=& \normal{\bb{z}_t}{\mu_{\phi}\left(\bb{x}_{\mathcal{T}_t}, y_t\right)}{\text{diag}\left(\sigma_{\phi}^2\left(\bb{x}_{\mathcal{T}_t}, y_t\right)\right)}. \label{eq:qz}
\end{eqnarray}
This combination of amortization and variational approximation results in the distribution $q_{\phi}\left(y_t | \bb{x}_{\mathcal{T}_t}\right)$ which can be interpreted as a classifier that can efficiently predict the discrete behavior state after training the model is complete. 

There has been much work on developing inference strategies for rSLDS models \citep{linderman2016rslds, linderman2017bayesian, nassar2018tree}. Our contribution here is the development of an efficient and flexible amortized variational inference strategy that incorporates both labeled and unlabeled data. We note that the proposed inference strategy leads to the definition of an approximate posterior $q_{\phi}\left(y_t | \bb{x}_{\mathcal{T}_t}\right)$. This approximate posterior -- which is trained on both labeled and unlabeled data -- also serves as a classifier, predicting the discrete action class $y_t$ from the observed data $\bb{x}_{1:T}$ using a neural network \citep{willetts2020semi}. Existing inference methods for rSLDS models do not use labeled data or an amortized inference network \citep{linderman2016rslds, linderman2017bayesian, nassar2018tree}. 
We use a TCN \citep{lea2016temporal,lea2017temporal} as our model backbone, as motivated in a previous section. This inference approach allows us to directly compare classifiers that are trained with labeled data only (i.e., supervised) or with both labeled and unlabeled data (i.e., semi-supervised). {Additionally, this approach improves both training and inference speeds over traditional fully unsupervised methods (Fig.~\ref{fig:time_chart}).}

\paragraph{ELBO with all labels observed}
Let us first consider the variational lower bound assuming all $y_t$ are observed (details in Section~\ref{app:elbo_rsnlds}). We will use $ \ \Tilde{} \ $ to denote that a variable is sampled from its approximate posterior using the reparameterization trick \citep{kingma2013auto}:
\begin{align} \label{eq:rslds_elbo_sup_f}
\mathcal{L}_l\left(\bb{x}_{1:T}, y_{1:T}\right) 
= & \sum_{t=1}^T {\log p\left(\bb{x}_t | \Tilde{\bb{z}}_t\right)} \nonumber \\
&- \KL\left[q\left(\bb{z}_1 | \bb{x}_{\mathcal{T}_1}, y_1\right) || \poz \right] + \log \poy
\nonumber \\
&- \sum_{t=2}^T {\KL\left[ \qtzgxsy || p\left(\bb{z}_t | \Tilde{\bb{z}}_{t-1}, y_t\right) \right]} \nonumber\\
& + \sum_{t=2}^T \log p\left(y_t | y_{t-1}, \Tilde{\bb{z}}_{t-1}\right).
\end{align}
Note that Eq.~\ref{eq:rslds_elbo_sup_f} does not explicitly contain a term that looks like a classifier; we will address this in a later section.

\paragraph{ELBO with all labels unobserved}
Next, let us derive the variational lower bound assuming all $y_t$ are unobserved (details in Section~\ref{app:elbo_rsnlds}).

\begin{align} \label{eq:rslds_elbo_unsup_marg_f}
\mathcal{L}_u\left(\bb{x}_{1:T}\right) = & 
\sum_{t=1}^T \sum_{k=1}^K  \Big( q\left(y_t = k | \bb{x}_{\mathcal{T}_t}\right) {\log p\left(\bb{x}_t | {\tilde{\bb{z}}}_t^k\right)} \Big)  \nonumber \\
&- \sum_{k=1}^{K} \Big( q\left(y_1 =k | \bb{x}_{\mathcal{T}_t}\right) {\KL\left[q\left(\bb{z}_1 | \bb{x}_{\mathcal{T}_1}, y_1 =k\right) || \poz \right]}\Big) - \KL \left[ q\left(y_1 | \bb{x}_{\mathcal{T}_1}\right) || \poy \right]\nonumber\\
&- \sum_{t=2}^T \sum_{k=1}^{K} \sum_{k'=1}^K \Big( q\left(y_t =k | \bb{x}_{\mathcal{T}_t}\right) q\left(y_{t-1} =k' | \bb{x}_{\mathcal{T}_{t-1}}\right)\nonumber\\
& \quad \quad{\KL\left[ {q\left(\bb{z}_t | \bb{x}_{\mathcal{T}_t}, y_t =k\right)} || {p\left(\bb{z}_t | \tilde{\bb{z}}_{t-1}^{k'},y_t = k\right)} \right]} \Big) \nonumber \\[1ex]
&- \sum_{t=2}^T \sum_{k=1}^{K} \Big( q\left(y_{t-1} =k | \bb{x}_{\mathcal{T}_{t-1}}\right){\KL\left[ \qtygxs || {p\left(y_t | y_{t-1} =k, \tilde{\bb{z}}_{t-1}^k\right)} \right]} \Big)
\end{align}

where $\Tilde{\bb{z}}_t^k \sim q\left(\bb{z}_t | \bb{x}_{\mathcal{T}_t}, y_t =k\right)$ and 
 $\Tilde{\bb{z}}_{t-1}^{k} \sim q\left(\bb{z}_{t-1} | \bb{x}_{\mathcal{T}_{t-1}}, y_{t-1} =k\right)$. A nice property of this lower bound is that each of the individual terms in the unlabeled ELBO (Eq.~\ref{eq:rslds_elbo_unsup_marg_f}) reduces to the corresponding term in the labeled ELBO (Eq.~\ref{eq:rslds_elbo_sup_f}) when the output of $q\left(y_{t} | \bb{x}_{\mathcal{T}_{t}}\right)$ is a one-hot vector (i.e. observed), which we exploit when computing the full semi-supervised ELBO (more below).

\paragraph{Classification loss} Note that Eq.~\ref{eq:rslds_elbo_unsup_marg_f} contains terms of the form $q\left(y_t | \bb{x}_{\mathcal{T}_t}\right)$, which maps observations to the discrete state. This part of the approximate posterior is a classifier that only appears in the unlabeled ELBO, and hence does not utilize labeled data for learning. Therefore, we can improve classification performance by including a cross entropy loss term for our model to learn the classes. We weight this loss with a hyperparameter $\alpha$, chosen through cross-validation.

\paragraph{Total loss function} The total loss function for the general semi-supervised case will be a combination of Eqs.~\ref{eq:rslds_elbo_sup_f} and \ref{eq:rslds_elbo_unsup_marg_f} depending on which subset of time points include observed discrete labels. In practice we compute $q\left(y_t | \bb{x}_{\mathcal{T}_t}\right)$ for all time points, then replace the probability vector with a one-hot vector when the discrete label is observed, and then compute Eq.~\ref{eq:rslds_elbo_unsup_marg_f}. In addition, we compute the classification loss for all observed discrete labels. See Section~\ref{app:elbo_rsnlds} for full details.

\subsection{Gaussian mixture deep generative model (GMDGM)}

For one of our baselines we implement the Gaussian mixture deep generative model (GMDGM) proposed in \cite{willetts2020semi}, which is similar to the \scubed~but without temporal dependencies (Fig.~\ref{fig:models_ablations}A). This is a modified version of the semi-supervised deep generative model proposed in \cite{kingma2014semi}.

\subsubsection{GMDGM model formulation}

The generative model is defined as
\begin{eqnarray}
p_{\theta}\left(\bb{x}, y, \bb{z}\right) &=& p_{\theta}\left(\bb{x} | \bb{z}\right)p_{\theta}\left(\bb{z} | y\right)p\left(y\right) \\
p_{\theta}\left(\bb{x} | \bb{z}\right) &=& \normal{\bb{x}}{g_{\theta}\left(\bb{z}\right)}{R} \\
p_{\theta}\left(\bb{z} | y\right) &=& \normal{\bb{z}}{f_{\theta}\left(y\right)}{\text{diag}\left(\sigma_{\theta}^2\left(y\right)\right)} \\
p\left(y\right) &=& \text{Cat}\left(y | \pi\right).
\end{eqnarray}
This model defines an explicit clustering mechanism by conditioning the latents $\bb{z}$ on the discrete label $y$, and defining a categorical prior over $y$. Therefore, even in the absence of labeled data, the GMDGM will attempt to cluster the data. 

\subsubsection{GMDGM inference and learning}
Like the \scubed~we perform approximate inference in the GMDGM model. Furthermore, we structure the approximate posteriors in a way that leads to a classifier $q_{\phi}\left(y | \bb{x}\right)$, which is trained on both labeled and unlabeled data. 
\begin{eqnarray}
q_{\phi}\left(y, \bb{z} | \bb{x}\right) &=& q_{\phi}\left(y | \bb{x}\right) q_{\phi}\left(\bb{z} | \bb{x}, y\right) \\
q_{\phi}\left(y | \bb{x}\right) &=& \text{Cat}\left(y | \pi_{\phi}\left(\bb{x}\right)\right) \\
q_{\phi}\left(\bb{z} | \bb{x}, y\right) &=& \normal{\bb{z}}{\mu_{\phi}\left(\bb{x}, y\right)}{\text{diag}\left(\sigma_{\phi}^2\left(\bb{x}, y\right)\right)}.
\end{eqnarray}
For the GMDGM-TCN model we use the same approximate posteriors as the \scubed~model, defined in Eqs.~\ref{eq:qyz}-\ref{eq:qz} (Fig.~\ref{fig:models_ablations}B).

The variational lower bound for labeled data is
\begin{eqnarray}
\mathcal{L}_l\left(\bb{x}, y\right) &=& \Eold_{q\left(\bb{z} | \bb{x}, y\right)}\left[\log \frac{p\left(\bb{x} | \bb{z}\right) p\left(\bb{z} | y\right) p\left(y\right)}{q\left(\bb{z} | \bb{x}, y\right)} \right] \nonumber \\
&=& \Eold_{q\left(\bb{z} | \bb{x}, y\right)}\left[\log p\left(\bb{x} | \bb{z}\right) \right] + \log p\left(y\right) - \KL[q\left(\bb{z} | \bb{x}, y \right) || p\left(\bb{z} | y\right)]  \nonumber \\ 
&\approx&  \log p\left(\bb{x} | \tilde{\bb{z}}\right) + \log p\left(y\right) - \KL[q\left(\bb{z} | \bb{x}, y \right) || p\left(\bb{z} | y\right)]
\end{eqnarray}

The variational lower bound for unlabeled data is
\begin{eqnarray}
\mathcal{L}_u\left(\bb{x}\right) &=& \Eold_{q\left(y, \bb{z} | \bb{x}\right)}\left[\log \frac{p\left(\bb{x} | \bb{z}\right) p\left(\bb{z} | y\right) p\left(y\right)}{q\left(y, \bb{z} | \bb{x}\right)} \right] \nonumber \\
&=& \Eold_{q\left(y, \bb{z} | \bb{x}\right)}\left[\log \frac{p\left(\bb{x} | \bb{z}\right) p\left(\bb{z} | y\right) p\left(y\right)}{q\left(y | \bb{x}\right)q\left(\bb{z} | \bb{x}, y\right)} \right] \nonumber \\
&=& \sum_y q\left(y | \bb{x}\right) \left[ \Eold_{q\left(\bb{z} | \bb{x}, y\right)}\left[\log \frac{p\left(\bb{x} | \bb{z}\right) p\left(\bb{z} | y\right) p\left(y\right)}{q\left(y | \bb{x}\right)q\left(\bb{z} | \bb{x}, y\right)} \right] \right] \nonumber \\
&=& \sum_y q\left(y | \bb{x}\right) \left[ \Eold_{q\left(\bb{z} | \bb{x}, y\right)}\left[\log \frac{p\left(\bb{x} | \bb{z}\right) p\left(\bb{z} | y\right) p\left(y\right)}{q\left(\bb{z} | \bb{x}, y\right)} \right] - \log q\left(y | \bb{x}\right) \right] \nonumber \\
&=& \sum_y q\left(y | \bb{x}\right) \left[\mathcal{L}_l\left(\bb{x}, y\right) - \log q\left(y | \bb{x}\right) \right] \nonumber \\
&=& \sum_y q\left(y | \bb{x}\right) \mathcal{L}_l\left(\bb{x}, y\right) - H\left(q\left(y | \bb{x}\right)\right) \nonumber \\
&=& \sum_{k=1}^K q\left(y=k | \bb{x}\right) \mathcal{L}_l\left(\bb{x}, k\right) - H\left(q\left(y | \bb{x}\right)\right).
\end{eqnarray}
As with the S$^3$LDS, note that the term $q\left(y | \bb{x}\right)$, which we will use as a final classifier, only appears in $\mathcal{L}_u\left(\bb{x}\right)$ and is therefore not trained with labeled data. We again address this by adding the cross entropy loss between the ground truth label and the distribution provided by $q\left(y | \bb{x}\right)$ to the lower bound of the labeled data, so that
\begin{eqnarray}
\mathcal{L} = \Eold_{\bb{x}^{\left(l\right)}, y^{\left(l\right)} \sim \datasetl}\left[\mathcal{L}_l\left(\bb{x}^{\left(l\right)}, y^{\left(l\right)}\right) + \alpha \log q\left(y^{\left(l\right)} | \bb{x}^{\left(l\right)}\right)\right] + \Eold_{\bb{x}^{\left(u\right)} \sim \datasetu}\left[\mathcal{L}_u\left(\bb{x}^{\left(u\right)}\right)\right],
\end{eqnarray}
where $\datasetl$ is the labeled dataset and $\datasetu$ is the unlabeled dataset.

\subsection{Model implementation details} \label{app:implementation}

\subsubsection{Temporal Convolutional Network backbone} 
We used the same dilated TCN backbone for the supervised model, GMDGM models, and S$^3$LDS. For all models and datasets we used 2 dilation blocks, 4 temporal lags per convolution layer, and 32 filters per convolution layer. Each dilation block consisted of a sequence of 2 sub-blocks (1D convolution layer $\rightarrow$ leaky ReLU nonlinearity $\rightarrow$ dropout with probability=0.10), as well as a residual connection between the input and output of the dilation block. The dilation of the convolutional filters starts with 1 for the first dilation block, then increases by a factor of 2 for each additional dilation block. This results in a larger temporal receptive field as the model gets deeper, allowing for learning of longer range dependencies \citep{yu2015multi}. For our specific parameters this results in outputs of the network at time $t$ being dependent on time points $\{t-24, \ldots, t, \ldots, t+24\}$, or in other words a 0.7 s receptive field for the fly data (70 Hz), a 1.96 s receptive field for the freely moving mouse data (25 Hz), and a 0.82 s receptive field for the head-fixed mouse and human data (both 60 Hz). We experimented with different values of these hyperparameters but found the results to be robust to specific choices. {Additionally, we use a weighted cross entropy loss function when training any model with labels, with class weights inversely proportional to the class frequency in the training data.}

\subsubsection{\scubed~model}
The \scubed~model used two TCN networks as described above: one for the approximate posterior of the discrete states $\qtygxs$, and one for the approximate posterior of the continuous latents $\qtzgxsyt$. We used a one-hidden-layer dense neural network for all the generative model nonlinearities in the \scubed~and S$^3$NLDS. The number of hidden units was always equal to the dimensionality of the continuous latent space. To determine this, we performed PCA on the observed data and chose the number of dimensions that explained 95\% of the variance.

\subsubsection{Model training} 
To train the supervised TCN, GMDGM, and \scubed~models we used the Adam optimizer \citep{kingma2014adam} with a learning rate of $0.0001$. Each batch contained 8 sequences of 1000 frames each. We trained the models for 500 epochs. For the semi-supervised models, we tested a range of possible $\alpha$s (the hyperparameter on the supervised classification loss). We chose $\alpha = 100$ for all of the datasets since that value generalized well across datasets. We anneal the two KL losses and $\log{y}$ (for the labeled data) by linearly increasing the weight from 0 to 1 over 100 epochs.

\subsection{Tree-based methods}

For the tree-based methods we used the same input features as the other models. We used a temporal window of features to match the input to the first layer of the TCN, i.e. the prediction at time $t$ is made from the concatenated behavioral features from $\{t-4, \ldots, t, \ldots, t+4\}$.

To train the random forest models we used the function \texttt{RandomForestClassifier} from the \texttt{sklearn} package with the following arguments: \texttt{n\_estimators=6000}, \texttt{max\_features=`sqrt'}, \texttt{criterion=`entropy'}, \texttt{min\_samples\_leaf=1}, \texttt{bootstrap=True} following \cite{goodwin2024simple}.

To train the XGBoost models we used the function \texttt{XGBClassifier} from the \texttt{xgboost} package with the following arguments: \texttt{n\_estimators=2000}, \texttt{max\_depth=3}, \texttt{learning\_rate=0.1}, \texttt{objective=`multi:softprob'}, \texttt{eval\_metric=`mlogloss'}, \texttt{tree\_method=`hist'}, \texttt{gamma=1}, \texttt{min\_child\_weight=1}, \texttt{subsample=0.8}, \texttt{colsample\_bytree=0.8} following \cite{segalin2021mouse}.

\subsection{Keypoint-MoSeq}
For keypoint-MoSeq, we use both the position features (Fig.~\ref{fig:results_kpm_other_m}) as well as the position-velocity features (Figs.~\ref{fig:results_kpm_fly}, \ref{fig:results_kpm_others})  as inputs for the head-fixed fly, freely moving mouse, and head-fixed mouse datasets. For the HuGaDB data, we use the IMU features. We used the \texttt{keypoint\_moseq} library to train the models. We initialized the model with ARHMM run for fifty iterations followed by training the full keypoint-MoSeq model for 500 additional iterations. We set \texttt{kappa=1e4} for all models.  
\clearpage
\newpage
\bibliography{bib}

\begin{thebibliography}{75}
\expandafter\ifx\csname natexlab\endcsname\relax\def\natexlab#1{#1}\fi
\expandafter\ifx\csname url\endcsname\relax
  \def\url#1{\texttt{#1}}\fi
\expandafter\ifx\csname urlprefix\endcsname\relax\def\urlprefix{URL: }\fi

\bibitem[{Ackerson and Fu(1970)}]{guy1970state}
Ackerson, G.~A. and Fu, K.-S. (1970) On state estimation in switching environments.
\newblock \textit{IEEE Transactions on Automatic Control}.

\bibitem[{Anderson and Perona(2014)}]{anderson2014toward}
Anderson, D.~J. and Perona, P. (2014) Toward a science of computational ethology.
\newblock \textit{Neuron}, \textbf{84}, 18--31.

\bibitem[{Azabou et~al.(2024)Azabou, Mendelson, Ahad, Sorokin, Thakoor, Urzay and Dyer}]{azabou2024relax}
Azabou, M., Mendelson, M., Ahad, N., Sorokin, M., Thakoor, S., Urzay, C. and Dyer, E. (2024) Relax, it doesn’t matter how you get there: A new self-supervised approach for multi-timescale behavior analysis.
\newblock \textit{Advances in Neural Information Processing Systems}, \textbf{36}.

\bibitem[{Batty et~al.(2019)Batty, Whiteway, Saxena, Biderman, Abe, Musall, Gillis, Markowitz, Churchland, Cunningham et~al.}]{batty2019behavenet}
Batty, E., Whiteway, M., Saxena, S., Biderman, D., Abe, T., Musall, S., Gillis, W., Markowitz, J., Churchland, A., Cunningham, J.~P. et~al. (2019) Behavenet: nonlinear embedding and bayesian neural decoding of behavioral videos.
\newblock \textit{Advances in Neural Information Processing Systems}, \textbf{32}.

\bibitem[{Berman et~al.(2014)Berman, Choi, Bialek and Shaevitz}]{berman2014mapping}
Berman, G.~J., Choi, D.~M., Bialek, W. and Shaevitz, J.~W. (2014) Mapping the stereotyped behaviour of freely moving fruit flies.
\newblock \textit{Journal of The Royal Society Interface}, \textbf{11}, 20140672.

\bibitem[{Biderman et~al.(2024)Biderman, Whiteway, Hurwitz, Greenspan, Lee, Vishnubhotla, Warren, Pedraja, Noone, Schartner et~al.}]{biderman2023lightning}
Biderman, D., Whiteway, M.~R., Hurwitz, C., Greenspan, N., Lee, R.~S., Vishnubhotla, A., Warren, R., Pedraja, F., Noone, D., Schartner, M.~M. et~al. (2024) Lightning pose: improved animal pose estimation via semi-supervised learning, bayesian ensembling and cloud-native open-source tools.
\newblock \textit{Nature Methods}, 1--13.

\bibitem[{Bohnslav et~al.(2021)Bohnslav, Wimalasena, Clausing, Yarmolinksy, Cruz, Chiappe, Orefice, Woolf and Harvey}]{Bohnslav2020deepetho}
Bohnslav, J.~P., Wimalasena, N.~K., Clausing, K.~J., Yarmolinksy, D., Cruz, T., Chiappe, E., Orefice, L.~L., Woolf, C.~J. and Harvey, C.~D. (2021) Deepethogram: a machine learning pipeline for supervised behavior classification from raw pixels.
\newblock \textit{eLife}.

\bibitem[{Breiman(2001)}]{breiman2001random}
Breiman, L. (2001) Random forests.
\newblock \textit{Machine Learning}, \textbf{45}, 5–32.

\bibitem[{Buchanan et~al.(2017)Buchanan, Lipschitz, Linderman and Paninski}]{buchanan2017quantifying}
Buchanan, E.~K., Lipschitz, A., Linderman, S.~W. and Paninski, L. (2017) Quantifying the behavioral dynamics of c. elegans with autoregressive hidden markov models.
\newblock In \textit{Workshop on Worm’s neural information processing at the 31st conference on neural information processing systems}.

\bibitem[{Chang and Athans(1978)}]{bing1978state}
Chang, C.-B. and Athans, M. (1978) State estimation for discrete systems with switching parameters.
\newblock \textit{IEEE Transactions on Aerospace and Electronic Systems}.

\bibitem[{Chen and Guestrin(2016)}]{chen2016xgboost}
Chen, T. and Guestrin, C. (2016) Xgboost: A scalable tree boosting system.
\newblock In \textit{Proceedings of the 22nd acm sigkdd international conference on knowledge discovery and data mining}, 785--794.

\bibitem[{Chereshnev and Kert{'e}sz-Farkas(2017)}]{chereshnev2017hugadb}
Chereshnev, R. and Kert{'e}sz-Farkas, A. (2017) Hugadb: Human gait database for activity recognition from wearable inertial sensor networks.
\newblock In \textit{International Conference on Analysis of Images, Social Networks and Texts}, 131--141. Springer.

\bibitem[{Costacurta et~al.(2022)Costacurta, Duncker, Sheffer, Gillis, Weinreb, Markowitz, Datta, Williams and Linderman}]{costacurta2022distinguishing}
Costacurta, J., Duncker, L., Sheffer, B., Gillis, W., Weinreb, C., Markowitz, J., Datta, S.~R., Williams, A. and Linderman, S. (2022) Distinguishing discrete and continuous behavioral variability using warped autoregressive hmms.
\newblock \textit{Advances in Neural Information Processing Systems}, \textbf{35}, 23838--23850.

\bibitem[{van Dam et~al.(2020)van Dam, Noldus and van Gerven}]{van2020deep}
van Dam, E.~A., Noldus, L.~P. and van Gerven, M.~A. (2020) Deep learning improves automated rodent behavior recognition within a specific experimental setup.
\newblock \textit{Journal of neuroscience methods}, \textbf{332}, 108536.

\bibitem[{Datta et~al.(2019)Datta, Anderson, Branson, Perona and Leifer}]{datta2019computational}
Datta, S.~R., Anderson, D.~J., Branson, K., Perona, P. and Leifer, A. (2019) Computational neuroethology: a call to action.
\newblock \textit{Neuron}, \textbf{104}, 11--24.

\bibitem[{Davidson et~al.(2021)Davidson, Buckermann, Steininger, Krause and Hotho}]{davidson2021semi}
Davidson, P., Buckermann, F., Steininger, M., Krause, A. and Hotho, A. (2021) Semi-unsupervised learning: An in-depth parameter analysis.
\newblock In \textit{German Conference on Artificial Intelligence (K{\"u}nstliche Intelligenz)}, 51--66. Springer.

\bibitem[{Dong et~al.(2020)Dong, Seybold, Murphy and Bui}]{dong2020collapsed}
Dong, Z., Seybold, B., Murphy, K. and Bui, H. (2020) Collapsed amortized variational inference for switching nonlinear dynamical systems.
\newblock In \textit{International Conference on Machine Learning}, 2638--2647. PMLR.

\bibitem[{Everett et~al.(2024)Everett, Norovich, Burke, Whiteway, Shih, Zhu, Paninski and Bendesky}]{everett2024coordination}
Everett, C.~P., Norovich, A.~L., Burke, J.~E., Whiteway, M.~R., Shih, P.-Y., Zhu, Y., Paninski, L. and Bendesky, A. (2024) Coordination and persistence of aggressive visual communication in siamese fighting fish.
\newblock \textit{bioRxiv}, 2024--04.

\bibitem[{Eyjolfsdottir et~al.(2016)Eyjolfsdottir, Branson, Yue and Perona}]{ey2016learn}
Eyjolfsdottir, E., Branson, K., Yue, Y. and Perona, P. (2016) Learning recurrent representations for hierarchical behavior modeling.
\newblock \textit{arXiv preprint arXiv:1611.00094}.

\bibitem[{Farha and Gall(2019)}]{farha2019ms}
Farha, Y.~A. and Gall, J. (2019) Ms-tcn: Multi-stage temporal convolutional network for action segmentation.
\newblock In \textit{Proceedings of the IEEE/CVF conference on computer vision and pattern recognition}, 3575--3584.

\bibitem[{Filtjens et~al.(2022)Filtjens, Vanrumste and Slaets}]{filtjens2022skeleton}
Filtjens, B., Vanrumste, B. and Slaets, P. (2022) Skeleton-based action segmentation with multi-stage spatial-temporal graph convolutional neural networks.
\newblock \textit{IEEE Transactions on Emerging Topics in Computing}.

\bibitem[{Fraccaro et~al.(2017)Fraccaro, Kamronn, Paquet and Winther}]{fraccaro2017disentangled}
Fraccaro, M., Kamronn, S., Paquet, U. and Winther, O. (2017) A disentangled recognition and nonlinear dynamics model for unsupervised learning.
\newblock \textit{Advances in neural information processing systems}, \textbf{30}.

\bibitem[{Gabriel et~al.(2022)Gabriel, Zeidler, Jin, Guo, Goodpaster, Kashay, Wu, Delaney, Cheung, DiFazio et~al.}]{gabriel2022behaviordepot}
Gabriel, C.~J., Zeidler, Z., Jin, B., Guo, C., Goodpaster, C.~M., Kashay, A.~Q., Wu, A., Delaney, M., Cheung, J., DiFazio, L.~E. et~al. (2022) Behaviordepot is a simple, flexible tool for automated behavioral detection based on markerless pose tracking.
\newblock \textit{Elife}, \textbf{11}, e74314.

\bibitem[{Glaser et~al.(2020)Glaser, Whiteway, Cunningham, Paninski and Linderman}]{glaser2020recurrent}
Glaser, J., Whiteway, M., Cunningham, J.~P., Paninski, L. and Linderman, S. (2020) Recurrent switching dynamical systems models for multiple interacting neural populations.
\newblock \textit{Advances in neural information processing systems}, \textbf{33}, 14867--14878.

\bibitem[{Gomez-Marin et~al.(2014)Gomez-Marin, Paton, Kampff, Costa and Mainen}]{gomez2014big}
Gomez-Marin, A., Paton, J.~J., Kampff, A.~R., Costa, R.~M. and Mainen, Z.~F. (2014) Big behavioral data: psychology, ethology and the foundations of neuroscience.
\newblock \textit{Nature neuroscience}, \textbf{17}, 1455--1462.

\bibitem[{Goodwin et~al.(2024)Goodwin, Choong, Hwang, Pitts, Bloom, Islam, Zhang, Szelenyi, Tong, Newman et~al.}]{goodwin2024simple}
Goodwin, N.~L., Choong, J.~J., Hwang, S., Pitts, K., Bloom, L., Islam, A., Zhang, Y.~Y., Szelenyi, E.~R., Tong, X., Newman, E.~L. et~al. (2024) Simple behavioral analysis (simba) as a platform for explainable machine learning in behavioral neuroscience.
\newblock \textit{Nature Neuroscience}, 1--14.

\bibitem[{Hamilton(1990)}]{ham1990time}
Hamilton, J.~D. (1990) Analysis of time series subject to changes in regime.
\newblock \textit{Journal of econometrics}.

\bibitem[{Hsu and Yttri(2021)}]{hsu2021bsoid}
Hsu, A.~I. and Yttri, E.~A. (2021) B-soid, an open-source unsupervised algorithm for identification and fast prediction of behaviors.
\newblock \textit{Nature communications}, \textbf{12}, 5188.

\bibitem[{{IBL} et~al.(2021){IBL}, Aguillon-Rodriguez, Angelaki, Bayer, Bonacchi, Carandini, Cazettes, Chapuis, Churchland, Dan et~al.}]{international2021standardized}
{IBL}, Aguillon-Rodriguez, V., Angelaki, D., Bayer, H., Bonacchi, N., Carandini, M., Cazettes, F., Chapuis, G., Churchland, A.~K., Dan, Y. et~al. (2021) Standardized and reproducible measurement of decision-making in mice.
\newblock \textit{Elife}, \textbf{10}, e63711.

\bibitem[{{IBL} et~al.(2023){IBL}, Benson, Benson, Birman, Bonacchi, Carandini, Catarino, Chapuis, Churchland, Dan et~al.}]{international2023brain}
{IBL}, Benson, B., Benson, J., Birman, D., Bonacchi, N., Carandini, M., Catarino, J.~A., Chapuis, G.~A., Churchland, A.~K., Dan, Y. et~al. (2023) A brain-wide map of neural activity during complex behaviour.
\newblock \textit{bioRxiv}, 2023--07.

\bibitem[{{IBL} et~al.(2022){IBL}, Birman, Bonacchi, Buchanan, Chapuis, Huntenburg, Meijer, Paninski, Schartner, Svoboda, Whiteway, Wells and Winter}]{ibl2022video}
{IBL}, Birman, D., Bonacchi, N., Buchanan, K., Chapuis, G., Huntenburg, J., Meijer, G., Paninski, L., Schartner, M., Svoboda, K., Whiteway, M., Wells, M. and Winter, O. (2022) Video hardware and software for the international brain laboratory.
\newblock \textit{figshare}.

\bibitem[{Jang et~al.(2017)Jang, Gu and Poole}]{jang2016categorical}
Jang, E., Gu, S. and Poole, B. (2017) Categorical reparameterization with gumbel-softmax.
\newblock \textit{International Conference on Learning Representations}.

\bibitem[{Jhuang et~al.(2010)Jhuang, Garrote, Yu, Khilnani, Poggio, Steele and Serre}]{jhuang2010automated}
Jhuang, H., Garrote, E., Yu, X., Khilnani, V., Poggio, T., Steele, A.~D. and Serre, T. (2010) Automated home-cage behavioural phenotyping of mice.
\newblock \textit{Nature communications}, \textbf{1}, 1--10.

\bibitem[{Johnson et~al.(2016)Johnson, Duvenaud, Wiltschko, Adams and Datta}]{johnson2016composing}
Johnson, M.~J., Duvenaud, D.~K., Wiltschko, A., Adams, R.~P. and Datta, S.~R. (2016) Composing graphical models with neural networks for structured representations and fast inference.
\newblock \textit{Advances in neural information processing systems}, \textbf{29}.

\bibitem[{Kabra et~al.(2013)Kabra, Robie, Rivera-Alba, Branson and Branson}]{kabra2013jaaba}
Kabra, M., Robie, A.~A., Rivera-Alba, M., Branson, S. and Branson, K. (2013) Jaaba: interactive machine learning for automatic annotation of animal behavior.
\newblock \textit{Nature methods}, \textbf{10}, 64.

\bibitem[{Karniol-Tambour et~al.(2024)Karniol-Tambour, Zoltowski, Diamanti, Pinto, Brody, Tank and Pillow}]{karniol2023modeling}
Karniol-Tambour, O., Zoltowski, D.~M., Diamanti, E.~M., Pinto, L., Brody, C.~D., Tank, D.~W. and Pillow, J.~W. (2024) Modeling state-dependent communication between brain regions with switching nonlinear dynamical systems.
\newblock In \textit{The Twelfth International Conference on Learning Representations}.

\bibitem[{Kim and Reiter(2017)}]{kim2017interpretable}
Kim, T.~S. and Reiter, A. (2017) Interpretable 3d human action analysis with temporal convolutional networks.
\newblock In \textit{2017 IEEE conference on computer vision and pattern recognition workshops (CVPRW)}, 1623--1631. IEEE.

\bibitem[{Kingma and Ba(2015)}]{kingma2014adam}
Kingma, D.~P. and Ba, J. (2015) Adam: A method for stochastic optimization.
\newblock \textit{International Conference on Learning Representations}.

\bibitem[{Kingma et~al.(2014)Kingma, Mohamed, Rezende and Welling}]{kingma2014semi}
Kingma, D.~P., Mohamed, S., Rezende, D.~J. and Welling, M. (2014) Semi-supervised learning with deep generative models.
\newblock In \textit{Advances in Neural Information Processing Systems}, 3581--3589.

\bibitem[{Kingma and Welling(2013)}]{kingma2013auto}
Kingma, D.~P. and Welling, M. (2013) Auto-encoding variational bayes.
\newblock \textit{arXiv preprint arXiv:1312.6114}.

\bibitem[{Kramida et~al.(2016)Kramida, Aloimonos, Parameshwara, Ferm{\"u}ller, Francis and Kanold}]{kramida2016automated}
Kramida, G., Aloimonos, Y., Parameshwara, C.~M., Ferm{\"u}ller, C., Francis, N.~A. and Kanold, P. (2016) Automated mouse behavior recognition using vgg features and lstm networks.
\newblock In \textit{Proc. Vis. Observ. Anal. Vertebrate Insect Behav. Workshop (VAIB)}, 1--3.

\bibitem[{Krishnan et~al.(2017)Krishnan, Shalit and Sontag}]{krishnan2017structured}
Krishnan, R., Shalit, U. and Sontag, D. (2017) Structured inference networks for nonlinear state space models.
\newblock In \textit{Proceedings of the AAAI Conference on Artificial Intelligence}, vol.~31.

\bibitem[{Krishnan et~al.(2015)Krishnan, Shalit and Sontag}]{krishnan2015deep}
Krishnan, R.~G., Shalit, U. and Sontag, D. (2015) Deep kalman filters.
\newblock \textit{arXiv preprint arXiv:1511.05121}.

\bibitem[{Lea et~al.(2017)Lea, Flynn, Vidal, Reiter and Hager}]{lea2017temporal}
Lea, C., Flynn, M.~D., Vidal, R., Reiter, A. and Hager, G.~D. (2017) Temporal convolutional networks for action segmentation and detection.
\newblock In \textit{proceedings of the IEEE Conference on Computer Vision and Pattern Recognition}, 156--165.

\bibitem[{Lea et~al.(2016)Lea, Vidal, Reiter and Hager}]{lea2016temporal}
Lea, C., Vidal, R., Reiter, A. and Hager, G.~D. (2016) Temporal convolutional networks: A unified approach to action segmentation.
\newblock In \textit{European Conference on Computer Vision}, 47--54. Springer.

\bibitem[{Lee et~al.(2023)Lee, Warrington, Glaser and Linderman}]{lee2023switching}
Lee, H.~D., Warrington, A., Glaser, J. and Linderman, S. (2023) Switching autoregressive low-rank tensor models.
\newblock \textit{Advances in Neural Information Processing Systems}, \textbf{36}, 57976--58010.

\bibitem[{Linderman et~al.(2017)Linderman, Johnson, Miller, Adams, Blei and Paninski}]{linderman2017bayesian}
Linderman, S., Johnson, M., Miller, A., Adams, R., Blei, D. and Paninski, L. (2017) Bayesian learning and inference in recurrent switching linear dynamical systems.
\newblock In \textit{Artificial Intelligence and Statistics}, 914--922. PMLR.

\bibitem[{Linderman et~al.(2016)Linderman, Miller, Adams, Blei, Paninski and Johnson}]{linderman2016rslds}
Linderman, S.~W., Miller, A.~C., Adams, R.~P., Blei, D.~M., Paninski, L. and Johnson, M.~J. (2016) Recurrent switching linear dynamical systems.

\bibitem[{Luxem et~al.(2022)Luxem, Mocellin, Fuhrmann, K{\"u}rsch, Miller, Palop, Remy and Bauer}]{luxem2022identifying}
Luxem, K., Mocellin, P., Fuhrmann, F., K{\"u}rsch, J., Miller, S.~R., Palop, J.~J., Remy, S. and Bauer, P. (2022) Identifying behavioral structure from deep variational embeddings of animal motion.
\newblock \textit{Communications Biology}, \textbf{5}, 1267.

\bibitem[{Maddison et~al.(2017)Maddison, Mnih and Teh}]{maddison2016concrete}
Maddison, C.~J., Mnih, A. and Teh, Y.~W. (2017) The concrete distribution: A continuous relaxation of discrete random variables.
\newblock \textit{International Conference on Learning Representations}.

\bibitem[{Markowitz et~al.(2018)Markowitz, Gillis, Beron, Neufeld, Robertson, Bhagat, Peterson, Peterson, Hyun, Linderman et~al.}]{markowitz2018striatum}
Markowitz, J.~E., Gillis, W.~F., Beron, C.~C., Neufeld, S.~Q., Robertson, K., Bhagat, N.~D., Peterson, R.~E., Peterson, E., Hyun, M., Linderman, S.~W. et~al. (2018) The striatum organizes 3d behavior via moment-to-moment action selection.
\newblock \textit{Cell}, \textbf{174}, 44--58.

\bibitem[{Markowitz et~al.(2023)Markowitz, Gillis, Jay, Wood, Harris, Cieszkowski, Scott, Brann, Koveal, Kula et~al.}]{markowitz2023spontaneous}
Markowitz, J.~E., Gillis, W.~F., Jay, M., Wood, J., Harris, R.~W., Cieszkowski, R., Scott, R., Brann, D., Koveal, D., Kula, T. et~al. (2023) Spontaneous behaviour is structured by reinforcement without explicit reward.
\newblock \textit{Nature}, \textbf{614}, 108--117.

\bibitem[{Mathis et~al.(2018)Mathis, Mamidanna, Cury, Abe, Murthy, Mathis and Bethge}]{mathis2018deeplabcut}
Mathis, A., Mamidanna, P., Cury, K.~M., Abe, T., Murthy, V.~N., Mathis, M.~W. and Bethge, M. (2018) Deeplabcut: markerless pose estimation of user-defined body parts with deep learning.
\newblock \textit{Nature neuroscience}, \textbf{21}, 1281--1289.

\bibitem[{McInnes et~al.(2018)McInnes, Healy, Saul and Großberger}]{mcinnes2018umap}
McInnes, L., Healy, J., Saul, N. and Großberger, L. (2018) Umap: Uniform manifold approximation and projection for dimension reduction.
\newblock \textit{arXiv preprint arXiv:1802.03426}, \textbf{3}, 861.

\bibitem[{Murari et~al.(2019)}]{murari2019recurrent}
Murari, K. et~al. (2019) Recurrent 3d convolutional network for rodent behavior recognition.
\newblock In \textit{ICASSP 2019-2019 IEEE International Conference on Acoustics, Speech and Signal Processing (ICASSP)}, 1174--1178. IEEE.

\bibitem[{Nassar et~al.(2018)Nassar, Linderman, Bugallo and Park}]{nassar2018tree}
Nassar, J., Linderman, S.~W., Bugallo, M. and Park, I.~M. (2018) Tree-structured recurrent switching linear dynamical systems for multi-scale modeling.
\newblock In \textit{International Conference on Learning Representations}.

\bibitem[{Nguyen et~al.(2019)Nguyen, Phan, Lumbanraja, Faisal, Abapihi, Purnama, Delimayanti, Mahmudah, Kubo and Satou}]{nguyen2019applying}
Nguyen, N.~G., Phan, D., Lumbanraja, F.~R., Faisal, M.~R., Abapihi, B., Purnama, B., Delimayanti, M.~K., Mahmudah, K.~R., Kubo, M. and Satou, K. (2019) Applying deep learning models to mouse behavior recognition.
\newblock \textit{Journal of Biomedical Science and Engineering}, \textbf{12}, 183--196.

\bibitem[{Pereira et~al.(2019)Pereira, Aldarondo, Willmore, Kislin, Wang, Murthy and Shaevitz}]{pereira2019fast}
Pereira, T.~D., Aldarondo, D.~E., Willmore, L., Kislin, M., Wang, S. S.-H., Murthy, M. and Shaevitz, J.~W. (2019) Fast animal pose estimation using deep neural networks.
\newblock \textit{Nature methods}, \textbf{16}, 117--125.

\bibitem[{Pereira et~al.(2020)Pereira, Shaevitz and Murthy}]{pereira2020quantifying}
Pereira, T.~D., Shaevitz, J.~W. and Murthy, M. (2020) Quantifying behavior to understand the brain.
\newblock \textit{Nature neuroscience}, 1--13.

\bibitem[{Ravbar et~al.(2019)Ravbar, Branson and Simpson}]{ravbar2019automatic}
Ravbar, P., Branson, K. and Simpson, J.~H. (2019) An automatic behavior recognition system classifies animal behaviors using movements and their temporal context.
\newblock \textit{Journal of neuroscience methods}, \textbf{326}, 108352.

\bibitem[{Rosenberg and Hirschberg(2007)}]{rosenberg2007v}
Rosenberg, A. and Hirschberg, J. (2007) V-measure: A conditional entropy-based external cluster evaluation measure.
\newblock In \textit{Proceedings of the 2007 joint conference on empirical methods in natural language processing and computational natural language learning (EMNLP-CoNLL)}, 410--420.

\bibitem[{Schaffer et~al.(2023)Schaffer, Mishra, Whiteway, Li, Vancura, Freedman, Patel, Voleti, Paninski, Hillman et~al.}]{schaffer2023spatial}
Schaffer, E.~S., Mishra, N., Whiteway, M.~R., Li, W., Vancura, M.~B., Freedman, J., Patel, K.~B., Voleti, V., Paninski, L., Hillman, E.~M. et~al. (2023) The spatial and temporal structure of neural activity across the fly brain.
\newblock \textit{Nature Communications}, \textbf{14}, 5572.

\bibitem[{Segalin et~al.(2021)Segalin, Williams, Karigo, Hui, Zelikowsky, Sun, Perona, Anderson and Kennedy}]{segalin2021mouse}
Segalin, C., Williams, J., Karigo, T., Hui, M., Zelikowsky, M., Sun, J.~J., Perona, P., Anderson, D.~J. and Kennedy, A. (2021) The mouse action recognition system (mars) software pipeline for automated analysis of social behaviors in mice.
\newblock \textit{Elife}, \textbf{10}, e63720.

\bibitem[{Sturman et~al.(2020)Sturman, von Ziegler, Schl{\"a}ppi, Akyol, Privitera, Slominski, Grimm, Thieren, Zerbi, Grewe and Bohacek}]{Sturman2020}
Sturman, O., von Ziegler, L., Schl{\"a}ppi, C., Akyol, F., Privitera, M., Slominski, D., Grimm, C., Thieren, L., Zerbi, V., Grewe, B. and Bohacek, J. (2020) Deep learning-based behavioral analysis reaches human accuracy and is capable of outperforming commercial solutions.
\newblock \textit{Neuropsychopharmacology}, \textbf{45}, 1942--1952.

\bibitem[{Sun et~al.(2021)Sun, Karigo, Chakraborty, Mohanty, Wild, Sun, Chen, Anderson, Perona, Yue et~al.}]{sun2021multi}
Sun, J.~J., Karigo, T., Chakraborty, D., Mohanty, S.~P., Wild, B., Sun, Q., Chen, C., Anderson, D.~J., Perona, P., Yue, Y. et~al. (2021) The multi-agent behavior dataset: Mouse dyadic social interactions.
\newblock \textit{Advances in neural information processing systems}, \textbf{2021}, 1.

\bibitem[{Sun et~al.(2020)Sun, Kennedy, Zhan, Anderson, Yue and Perona}]{sun2020task}
Sun, J.~J., Kennedy, A., Zhan, E., Anderson, D.~J., Yue, Y. and Perona, P. (2020) Task programming: Learning data efficient behavior representations.

\bibitem[{Sun et~al.(2024)Sun, Zhou, Zhao, Yuan, Seybold, Hendon, Schroff, Ross, Adam, Hu et~al.}]{sun2024video}
Sun, J.~J., Zhou, H., Zhao, L., Yuan, L., Seybold, B., Hendon, D., Schroff, F., Ross, D.~A., Adam, H., Hu, B. et~al. (2024) Video foundation models for animal behavior analysis.
\newblock \textit{bioRxiv}, 2024--07.

\bibitem[{Tucker et~al.(2017)Tucker, Mnih, Maddison, Lawson and Sohl-Dickstein}]{tucker2017rebar}
Tucker, G., Mnih, A., Maddison, C.~J., Lawson, J. and Sohl-Dickstein, J. (2017) Rebar: Low-variance, unbiased gradient estimates for discrete latent variable models.
\newblock \textit{Advances in Neural Information Processing Systems}, \textbf{30}.

\bibitem[{Weinreb et~al.(2024)Weinreb, Pearl, Lin, Osman, Zhang et~al.}]{weinreb2023keypoint}
Weinreb, C., Pearl, J.~E., Lin, S., Osman, M. A.~M., Zhang, L. et~al. (2024) Keypoint-moseq: parsing behavior by linking point tracking to pose dynamics.
\newblock \textit{Nature Methods}, \textbf{21}, 1329--1339.

\bibitem[{Whiteway et~al.(2021)Whiteway, Biderman, Friedman, Dipoppa, Buchanan, Wu, Zhou, Bonacchi, Miska, Noel et~al.}]{whiteway2021partitioning}
Whiteway, M.~R., Biderman, D., Friedman, Y., Dipoppa, M., Buchanan, E.~K., Wu, A., Zhou, J., Bonacchi, N., Miska, N.~J., Noel, J.-P. et~al. (2021) Partitioning variability in animal behavioral videos using semi-supervised variational autoencoders.
\newblock \textit{PLoS computational biology}, \textbf{17}, e1009439.

\bibitem[{Willetts et~al.(2020)Willetts, Roberts and Holmes}]{willetts2020semi}
Willetts, M., Roberts, S. and Holmes, C. (2020) Semi-unsupervised learning: Clustering and classifying using ultra-sparse labels.
\newblock In \textit{2020 IEEE International Conference on Big Data (Big Data)}, 5286--5295. IEEE.

\bibitem[{Wiltschko et~al.(2015)Wiltschko, Johnson, Iurilli, Peterson, Katon, Pashkovski, Abraira, Adams and Datta}]{wiltschko2015mapping}
Wiltschko, A.~B., Johnson, M.~J., Iurilli, G., Peterson, R.~E., Katon, J.~M., Pashkovski, S.~L., Abraira, V.~E., Adams, R.~P. and Datta, S.~R. (2015) Mapping sub-second structure in mouse behavior.
\newblock \textit{Neuron}, \textbf{88}, 1121--1135.

\bibitem[{Wiltschko et~al.(2020)Wiltschko, Tsukahara, Zeine, Anyoha, Gillis, Markowitz, Peterson, Katon, Johnson and Datta}]{wilt2020pharm}
Wiltschko, A.~B., Tsukahara, T., Zeine, A., Anyoha, R., Gillis, W.~F., Markowitz, J.~E., Peterson, R.~E., Katon, J., Johnson, M.~J. and Datta, S.~R. (2020) Revealing the structure of pharmacobehavioral space through motion sequencing.
\newblock \textit{Nature Neuroscience}, \textbf{23}.

\bibitem[{Yu and Koltun(2016)}]{yu2015multi}
Yu, F. and Koltun, V. (2016) Multi-scale context aggregation by dilated convolutions.
\newblock \textit{International Conference on Learning Representations}.

\bibitem[{von Ziegler et~al.(2021)von Ziegler, Sturman and Bohacek}]{von2021big}
von Ziegler, L., Sturman, O. and Bohacek, J. (2021) Big behavior: challenges and opportunities in a new era of deep behavior profiling.
\newblock \textit{Neuropsychopharmacology}, \textbf{46}, 33--44.

\end{thebibliography}


\clearpage
\newpage
\section{Supplementary figures} \label{sec:supp_figs}

\setcounter{figure}{0}
\renewcommand{\thefigure}{S\arabic{figure}}

\makeatletter
\setlength{\@fptop}{0pt}
\makeatother

\begin{figure}[h]
\centering
\includegraphics[width=\linewidth]{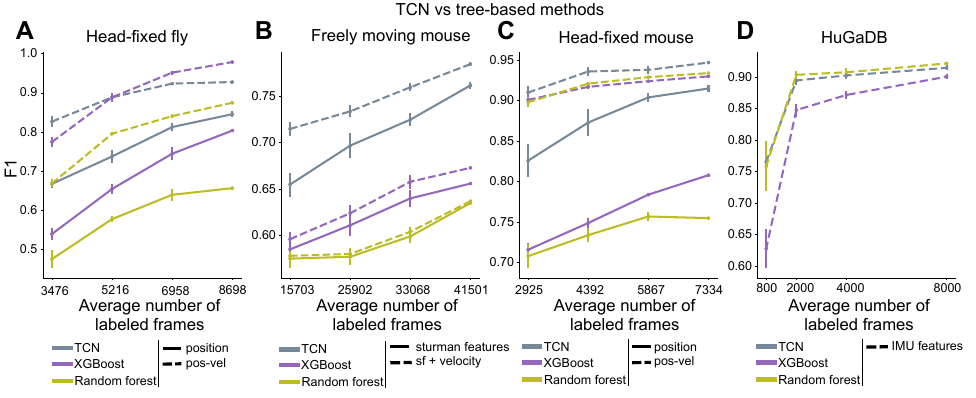}
\vspace{-0.05in}
\caption{\textbf{Temporal convolutional networks outperform tree-based methods in supervised classification.} We compare the TCN network to XGBoost and random forest models. We compare models trained on position features (solid lines), as well as on models trained on position-velocity features (dashed lines). For the position feature models, the TCN is the best model on all three datasets that use markers (\textbf{A}, \textbf{B}, \textbf{C}). For the position-velocity features, the TCN is the best model for the freely moving mouse and head-fixed mouse datasets (\textbf{B}, \textbf{C}). The TCN and XGBoost models perform similarly for the head-fixed fly dataset (\textbf{A}), and for the HuGaDB dataset, the TCN and random forest perform similarly (\textbf{D}). Given that the TCN performs better or similarly across all four datasets, we select this as the backbone model throughout the rest of the model comparisons in the paper.
}
\vspace{-0.05in}
 \label{fig:results_super_baselines}
\end{figure}
\vspace{1in}

\begin{figure}[h]
\centering
\includegraphics[width=\linewidth]{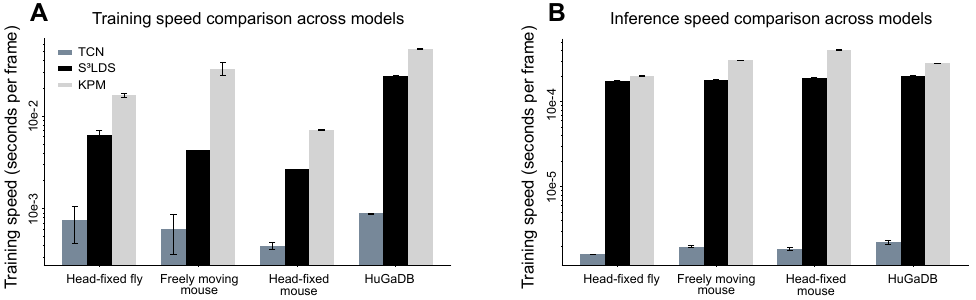}
\vspace{-0.05in}
\caption{{\textbf{Training and inference speed for different model types.} For all datasets, the TCN has the fastest training (\textbf{A}) and inference (\textbf{B}) times, followed by S$^3$LDS, and finally keypoint-MoSeq. Bars represent mean value over 5 random seeds, error bars show s.e.m.}}
\vspace{-0.05in}
 \label{fig:time_chart}
\end{figure}
\clearpage
\newpage

\begin{figure}[t]
\centering
\includegraphics[width=\linewidth]{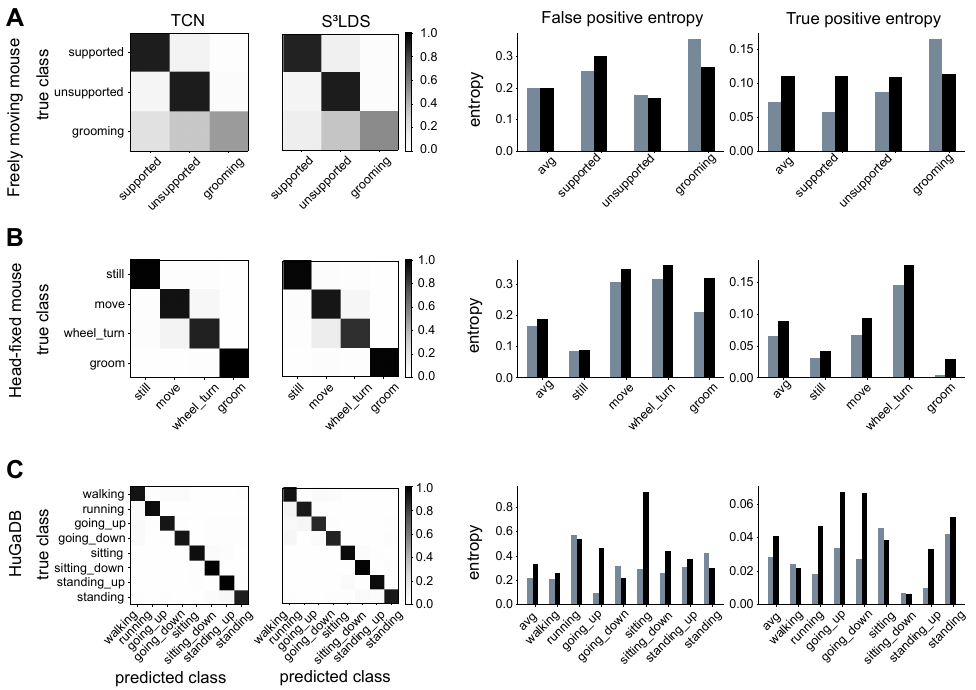}
\vspace{-0.05in}
\caption{\textbf{Semi-supervised models have less confident predictions than their supervised counterparts.}
The first two columns show the confusion matrices for the TCN and \scubed~models, respectively. The latter two columns display the average entropy of the false positives (left) and true positives (right) for both models. In both cases entropies are larger for the \scubed~model, indicating higher uncertainty in the state distributions. All results are from the models trained on position-velocity features. 
}
\vspace{-0.05in}
 \label{fig:results_super_others_detail}
\end{figure}
\clearpage
\newpage

\begin{figure}[t]
\centering
\includegraphics[width=\linewidth]{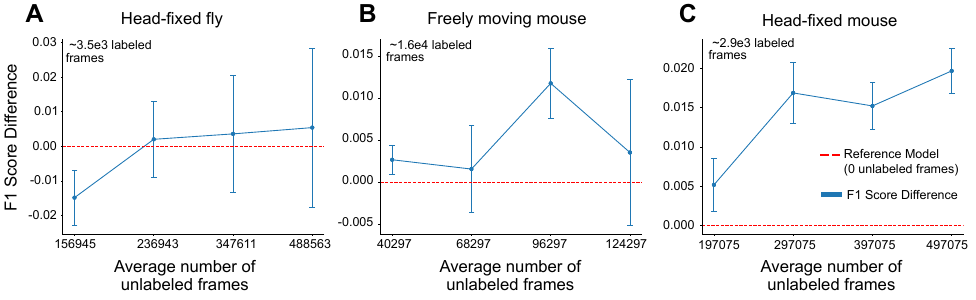}
\vspace{-0.05in}
\caption{{\textbf{Performance of \scubed \ when increasing the number of unlabeled frames: position features.}
F1 score differences comparing a \scubed \ trained without any unlabeled frames vs. adding different numbers of unlabeled frames, for models trained with position features. The F1 differences are averaged five models each trained on a different subset of data, with the error bar showing the standard deviation over all five sets of model differences. For all of the datasets, there is a positive trend between the number of unlabeled frames and the corresponding F1 scores.  
\textbf{A}: results on the head-fixed fly dataset. \textbf{B}: results on the freely moving mouse dataset. \textbf{C}: results on the HuGaDB data.
}}
\vspace{-0.05in}
 \label{fig:unlabeled_m}
\end{figure}

\begin{figure}[t]
\centering
\includegraphics[width=\linewidth]{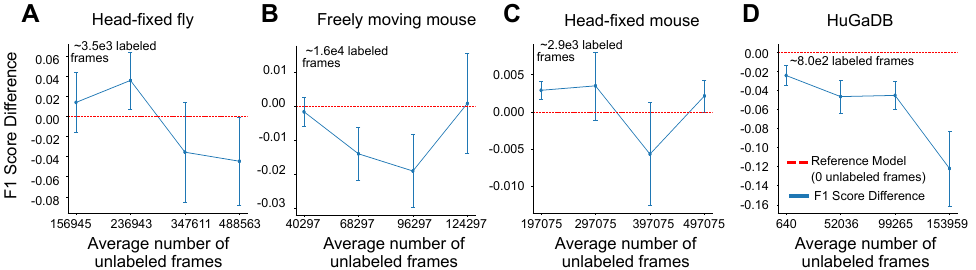}
\vspace{-0.05in}
\caption{{\textbf{Performance of \scubed \ when increasing the number of unlabeled frames: position-velocity features.}
F1 score differences comparing a \scubed \ trained without any unlabeled frames vs. adding different numbers of unlabeled frames, for models trained with position-velocity features. For all datasets, there is negative trend between the number of unlabeled frames and the corresponding F1 scores. This suggests that the semi-supervised \scubed \ is not a good fit for modeling these more complex features.
Conventions as in Fig.~\ref{fig:unlabeled_m}. 
}}
\vspace{-0.05in}
 \label{fig:unlabeled_pv}
\end{figure}

\begin{figure}[p!]
\centering
\includegraphics[width=\linewidth]{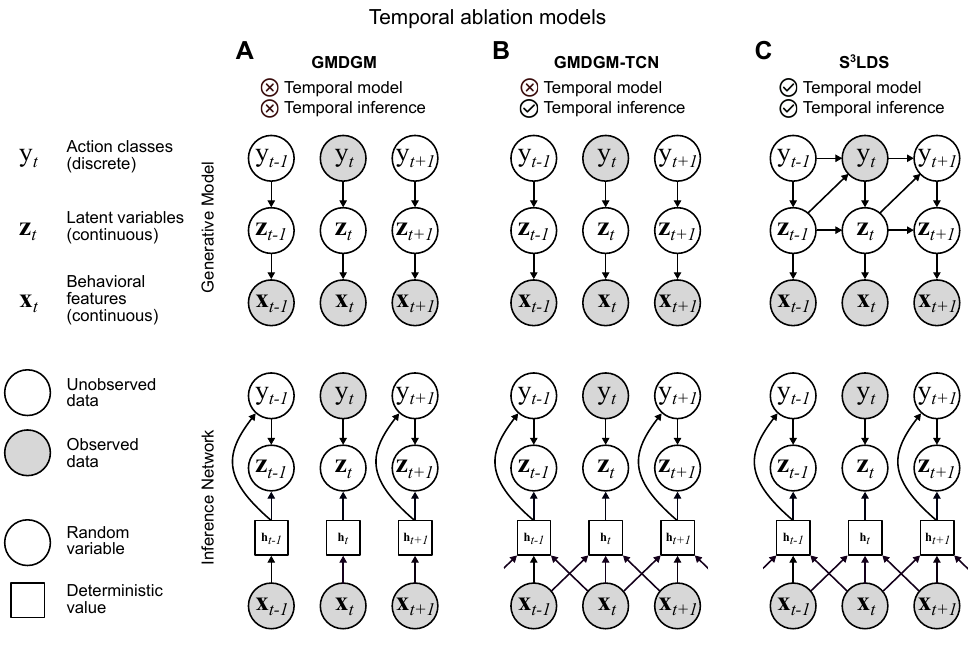}
\vspace{-0.05in}
\caption{\textbf{\scubed~model ablations.} 
\textbf{A}: Graphical model for the Gaussian Mixture Deep Generative Model (GMDGM) \citep{willetts2020semi}. The GMDGM does not contain temporal dependencies in the generative model (top) or the inference network (bottom).
\textbf{B}: The GMDGM-TCN, which does not contain temporal dependencies in the generative model (top) but does in the inference model (bottom), where we use a window of observed behavioral features for state prediction.  
\textbf{C}: The \scubed~contains temporal dependencies in both the generative model (top) and inference network (bottom). Same schematic as Fig.~\ref{fig:models}C.
}
\vspace{-0.05in}
 \label{fig:models_ablations}
\end{figure}
\clearpage
\newpage

\begin{figure}[t]
\centering
\includegraphics[width=\linewidth]{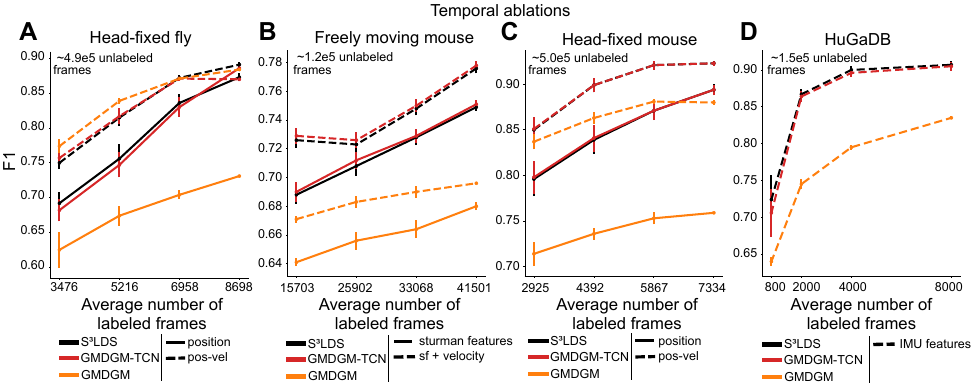}
\vspace{-0.05in}
\caption{\textbf{The role of temporal information in semi-supervised model performance.}
Temporal ablations comparing \scubed~(temporal inference and generative model), GMDGM-TCN (temporal inference only), and GMDGM (non-temporal) models. The non-temporal model performs much worse than the other two across all datasets, especially when using non-temporal features.
}
\vspace{-0.05in}
 \label{fig:results_super_ablations}
\end{figure}

\begin{figure}[t]
\centering
\includegraphics[width=\linewidth]{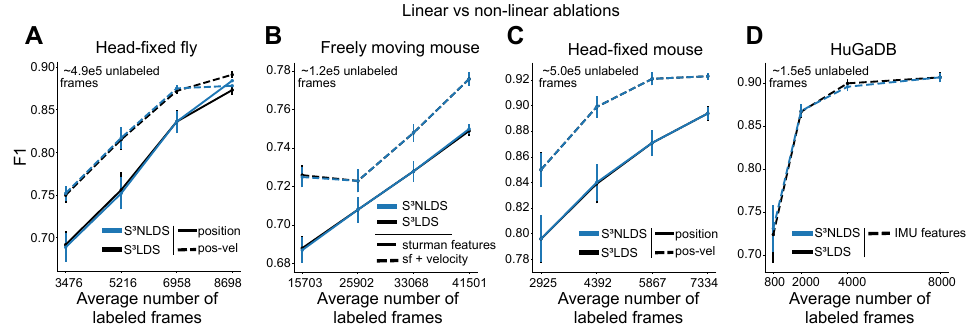}
\vspace{-0.05in}
\caption{\textbf{The role of nonlinearities in semi-supervised model performance.}
We compare our \scubed~model to a model that uses one-hidden-layer dense neural networks for the dynamics and recurrent transitions (S$^3$NLDS). The nonlinearities do not improve model performance on any dataset.
}
\vspace{-0.05in}
 \label{fig:results_super_nonlinearities}
\end{figure}
\clearpage
\newpage


\begin{figure}[t]
\centering
\includegraphics[width=\linewidth]{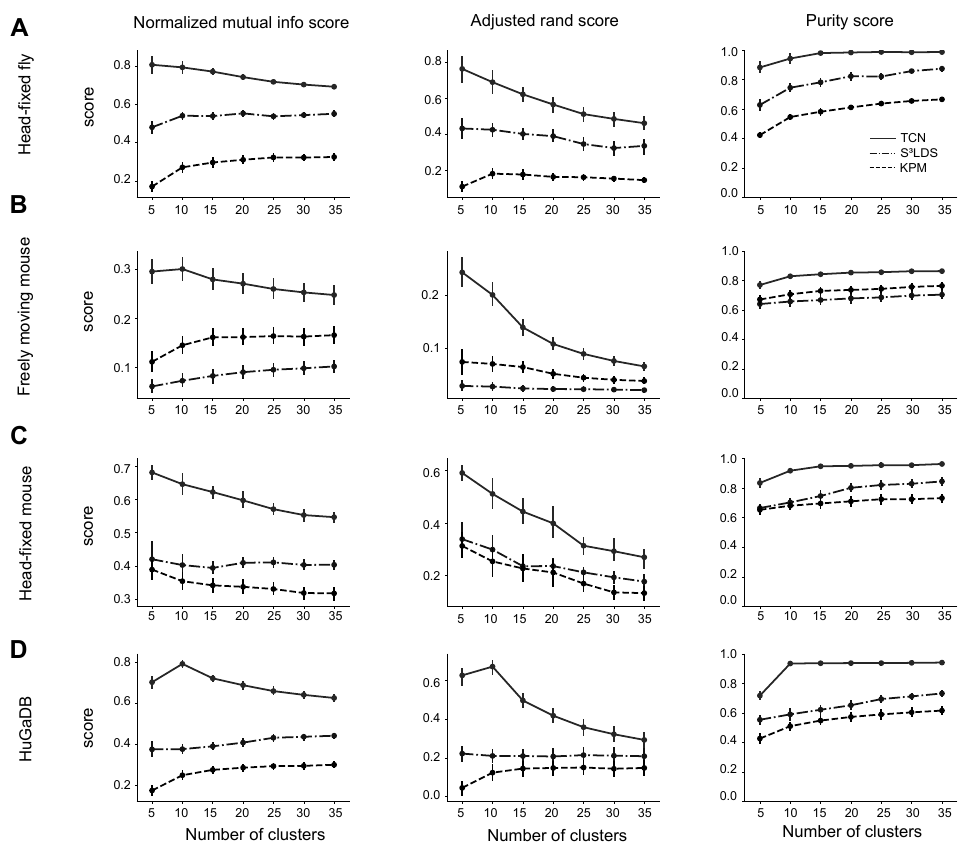}
\vspace{-0.05in}
\caption{\textbf{The latent spaces of models are shaped by the extent to which labels are used in training: position-velocity features.}
Cluster metrics for models trained with position-velocity features, which complement the cluster homogeneity score in Figs.~\ref{fig:results_kpm_fly},~\ref{fig:results_kpm_others}. The addition of hand labels produces more homogeneous clusters in the models’ latent spaces. 
}
\vspace{-0.05in}
 \label{fig:results_kpm_other_metrics_pv}
\end{figure}
\clearpage
\newpage

\begin{figure}[t]
\centering
\includegraphics[width=\linewidth]{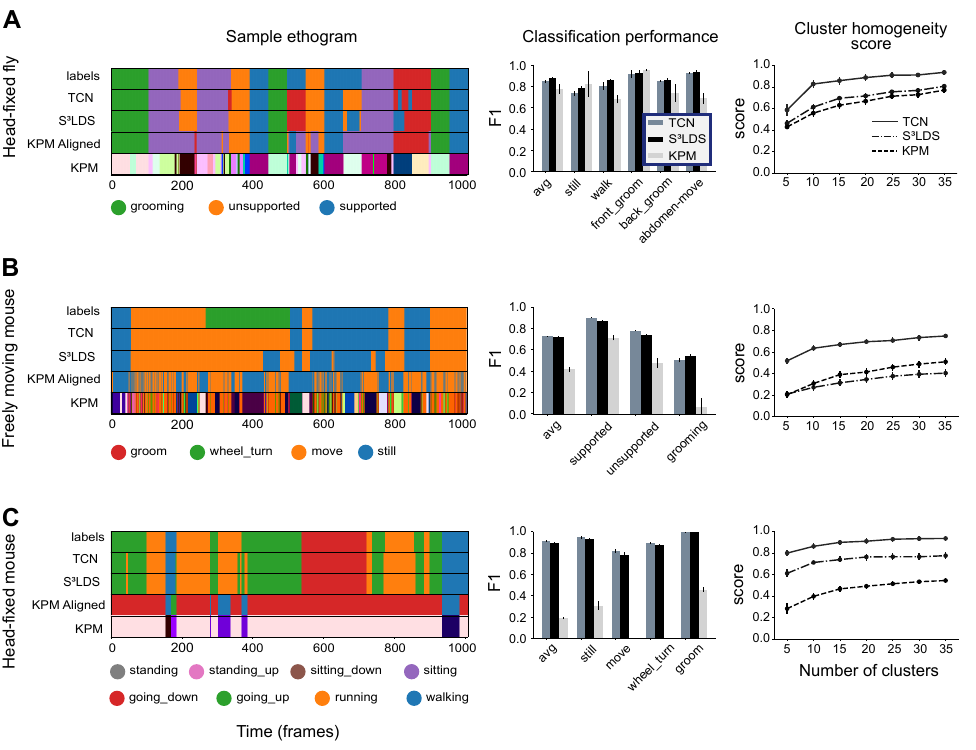}
\vspace{-0.05in}
\caption{\textbf{Keypoint-MoSeq performance on non-fly datasets: position features.}
Comparisons between the models using only position features. Conventions as in Fig.~\ref{fig:results_kpm_others}. We find the TCN, which is purely supervised, achieves the highest alignment of the latent space with the ground truth labels as measured by the cluster homogeneity score.
}
\vspace{-0.05in}
 \label{fig:results_kpm_other_m}
\end{figure}

\begin{figure}[t]
\centering
\includegraphics[width=\linewidth]{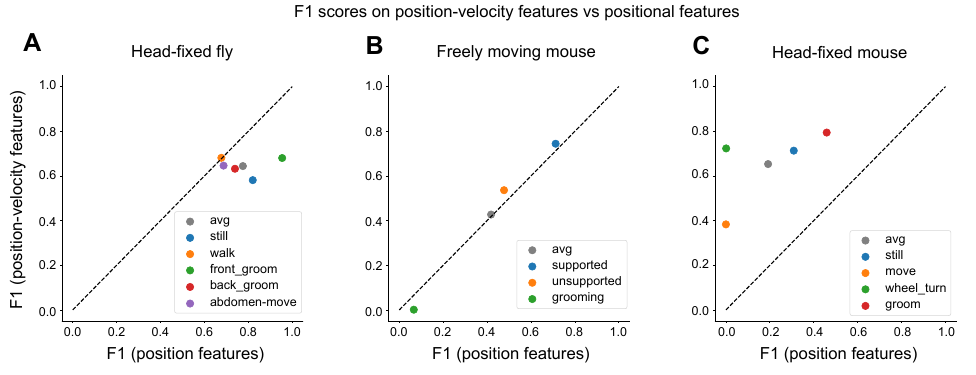}
\vspace{-0.05in}
\caption{\textbf{Keypoint-MoSeq performance: position-velocity features vs position features.}
Comparisons between the keypoint-Moseq models trained on position-velocity features vs position features. We find that keypoint-Moseq performs best with the position features on the head-fixed fly dataset, and the position-velocity features on the two mouse datasets. 
}
\vspace{-0.05in}
 \label{fig:results_kpm_other_m_f1}
\end{figure}
\clearpage
\newpage

\begin{figure}[t]
\centering
\includegraphics[width=\linewidth]{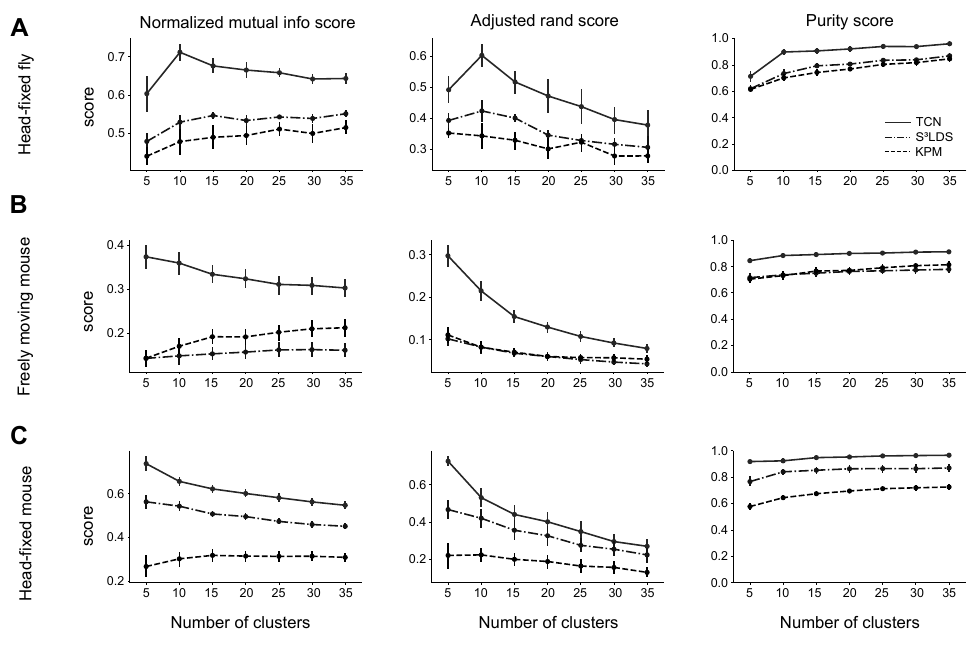}
\vspace{-0.05in}
\caption{\textbf{Cluster metrics for model latent spaces: position features.}
Cluster metrics for models trained with position features. Conventions as in Fig.~\ref{fig:results_kpm_other_metrics_pv}. 
}
\vspace{-0.05in}
 \label{fig:results_kpm_other_metrics_m}
\end{figure}

\clearpage
\newpage


\section{Supplementary tables} \label{sec:supp_tables}
\setcounter{table}{0}
\renewcommand{\thetable}{S\arabic{table}}

{\renewcommand{\arraystretch}{1.2}
\begin{table*}[h!]
\resizebox{\textwidth}{!}{
\centering{
\begin{tabular}{r  c  c  c  c  c  c}
\hline
Experiment ID & Total frames & Walk & Still & 
Front groom & Back groom & Abdomen move\\
\hline
\textbf{Train} \ \ \ \ \ 2019\_08\_07\_fly2 & 50000 & 300 & 300 & 100 & 300 & 591\\
2019\_08\_08\_fly1 & 94960 & 300 & 300 & 300 & 150 & 594\\
2019\_08\_20\_fly2 & 65108 & 300 & 300 & 300 & 300 & 1067\\
2019\_10\_10\_fly3 & 141042 & 300 & 300 & 300 & 300 & 478\\
2019\_10\_14\_fly3 & 140929 & 300 & 300 & 300 & 300 & 318\\
\hline
\textbf{Test} \ \ \ \ \ 2019\_06\_26\_fly2 & 45000 & 300 & 300 & 300 & 300 & 706\\
2019\_08\_14\_fly1 & 124144 & 300 & 300 & 300 & 300 & 693\\
2019\_08\_20\_fly3 & 73055 & 300 & 300 & 300 & 300 & 405\\
2019\_10\_14\_fly2 & 139925 & 300 & 300 & 300 & 300 & 101\\
2019\_10\_21\_fly1 & 142554 & 300 & 300 & 300 & 300 & 0\\
\hline
\textbf{Totals} & 1016717 & 3000 & 3000 & 2800 & 2850 & 4953\\
\hline
\end{tabular}
}
}
\caption{The number of labeled frames per behavior for the head-fixed fly dataset.}
\label{tab:labels-fly}
\end{table*}
}

\vspace{0.5in}

{\renewcommand{\arraystretch}{1.2}
\begin{table*}[h!]
\centering{
\begin{tabular}{c r  c  c  c  c}
\hline
& Experiment ID & Total frames & Supported rear & Unsupported rear & Groom \\
\hline
\textbf{Train}
& OFT\_5 & 14000 & 3267 & 1208 & 387 \\
& OFT\_6 & 14000 & 2424 & 1090 & 428 \\
& OFT\_11 & 14000 & 1900 & 1036 & 117 \\
& OFT\_12 & 14000 & 2354 & 2213 & 202 \\
& OFT\_14 & 14000 & 3165 & 2071 & 140 \\
& OFT\_15 & 14000 & 2038 & 1433 & 154 \\
& OFT\_16 & 14000 & 1773 & 826 & 493 \\
& OFT\_23 & 14000 & 1708 & 263 & 396 \\
& OFT\_24 & 14000 & 2130 & 2312 & 298 \\
& OFT\_38 & 14000 & 2564 & 2598 & 513 \\
\hline
\textbf{Test}
& OFT\_39 & 14000 & 2159 & 2062 & 570 \\
& OFT\_41 & 14000 & 2381 & 1479 & 178 \\
& OFT\_43 & 14000 & 3642 & 338 & 125 \\
& OFT\_44 & 14000 & 2314 & 1069 & 139 \\
& OFT\_49 & 14000 & 2638 & 1120 & 142 \\
& OFT\_50 & 14000 & 1550 & 873 & 386 \\
& OFT\_51 & 14000 & 1799 & 782 & 442 \\
& OFT\_52 & 14000 & 2949 & 1605 & 547 \\
& OFT\_54 & 14000 & 2587 & 2012 & 367 \\
& OFT\_58 & 14000 & 2468 & 1858 & 577 \\
\hline
& \textbf{Totals} & 280000 & 47810 & 28248 & 6601\\
\hline
\end{tabular}
}
\caption{The number of labeled frames per behavior for the freely moving mouse dataset.}
\label{tab:labels-mouse}
\end{table*}
}

\vspace{0.5in}

{\renewcommand{\arraystretch}{1.2}
\begin{table*}[h!]
\resizebox{\textwidth}{!}{
\centering{
\begin{tabular}{c r  c  c  c  c c}
\hline
& Experiment ID & Total frames & Still & Move & Wheel turn & Groom \\

\hline
\textbf{Train}
& danlab\_DY\_009\_2020-02-27-001 & 100000 & 497 & 316 & 374 & 300 \\
& danlab\_DY\_018\_2020-10-15-001 & 100000 & 400 & 268 & 241 & 580 \\
& hoferlab\_SWC\_061\_2020-11-23-001 & 100000 & 320 & 261 & 494 & 265 \\
& mrsicflogellab\_SWC\_058\_2020-12-11-001 & 100000 & 433 & 357 & 358 & 298 \\
& wittenlab\_ibl\_witten\_26\_2021-01-27-002 & 100000 & 624 & 235 & 353 & 360 \\
\hline
\textbf{Test}
& churchlandlab\_CSHL045\_2020-02-27-001 & 100000 & 306 & 368 & 522 & 469 \\
& cortexlab\_KS020\_2020-02-06-001 & 100000 & 378 & 340 & 322 & 208 \\
& hoferlab\_SWC\_043\_2020-09-15-001 & 100000 & 469 & 363 & 306 & 513 \\
& mrsicflogellab\_SWC\_052\_2020-10-22-001 & 100000 & 338 & 179 & 494 & 242 \\
& wittenlab\_ibl\_witten\_27\_2021-01-21-001 & 100000 & 335 & 35 & 536 & 50 \\
\hline
& \textbf{Totals} & 1000000 & 4100 & 2722 & 4000 & 3285 \\
\hline
\end{tabular}
}
}
\caption{The number of labeled frames per behavior for the IBL head-fixed mouse dataset.}
\label{tab:labels-ibl}
\end{table*}
}

\vspace{0.5in}

{\renewcommand{\arraystretch}{1.2}
\begin{table*}[h!]
\resizebox{\textwidth}{!}{
\centering{
\begin{tabular}{c r  c  c  c  c  c  c  c  c  c}
\hline
& Experiment ID & Total frames & Walking & Running & Going up & Going down & Sitting & Sitting down & Standing up & Standing \\
\hline
\textbf{Train}
& subject 05 & 56199 & 23034 & 2154 & 9820 & 10246 & 4215 & 834 &  1052 & 4844 \\
& subject 07 & 53571 & 23533& 4111& 5767& 6460& 5000& 1164& 1273& 6263\\
& subject 09 & 40066& 12254& 3712& 7374& 5954& 3278& 1038& 1227& 5229\\
& subject 12 & 52005& 14343& 5796& 7683& 7805& 6843& 908& 1085& 7542\\
& subject 14 & 55366& 23365& 5529& 8829& 6199& 4771& 920& 900& 4853\\
\hline

\textbf{Test}
& subject 06 & 67747& 33386& 11334& 5367& 6214& 3153& 922& 1008& 6363\\
& subject 08 & 49093& 17327& 4081& 7138& 6262& 4710& 1101& 1509& 6965\\
& subject 11 & 45137& 16202& 3727& 7768& 5797& 4471& 1257& 1329& 4586\\
& subject 13 & 52053& 17971& 5374& 10969& 6942& 3896& 1095& 1314& 4492\\
& subject 17 & 45547& 12579& 1862& 8723& 6130& 6867& 1276& 1235& 6875\\
\hline
& \textbf{Totals} & 516784& 193994& 47680& 79438& 68009& 47204& 10515& 11932& 58012\\
\hline
\end{tabular}
}
}

\caption{The number of labeled frames per behavior for the human gait database.}
\label{tab:labels-hugadb}
\end{table*}
}


\clearpage
\newpage


\section{Variational lower bound of \scubed~model} \label{app:elbo_rsnlds}
\subsection{All labels observed}
We first derive the variational lower bound, assuming all $y_t$ are observed. For clarity we drop the dependence on $\theta$ and $\phi$.
\begin{eqnarray} \label{eq:rslds_elbo_sup_1}
\log p\left(\bb{x}_{1:T}, y_{1:T}\right) & \geq & \int_{\bb{z}_{1:T}} q\left(\bb{z}_{1:T} | \bb{x}_{1:T}, y_{1:T}\right) \log \frac{p\left(\bb{x}_{1:T} | \bb{z}_{1:T}\right)p\left(\bb{z}_{1:T}, y_{1:T}\right)}{q\left(\bb{z}_{1:T} | \bb{x}_{1:T}, y_{1:T}\right)} \nonumber\\
&=& \E{q\left(\bb{z}_{1:T} | \bb{x}_{1:T}, y_{1:T}\right)}{\log p\left(\bb{x}_{1:T} | \bb{z}_{1:T}\right)} + \E{q\left(\bb{z}_{1:T} | \bb{x}_{1:T}, y_{1:T}\right)}{\log \frac{p\left(\bb{z}_{1:T}, y_{1:T}\right)}{q\left(\bb{z}_{1:T} | \bb{x}_{1:T}, y_{1:T}\right)}}. \nonumber
\end{eqnarray}
For the first term, recall that both $q\left(\bb{z}_{1:T} | \bb{x}_{1:T}, y_{1:T}\right)$ and $p\left(\bb{x}_{1:T} | \bb{z}_{1:T}\right)$ factorize across time:
\begin{eqnarray}
\E{q\left(\bb{z}_{1:T} | \bb{x}_{1:T}, y_{1:T}\right)}{\log p\left(\bb{x}_{1:T} | \bb{z}_{1:T}\right)} &=& \int_{\bb{z}_{1:T}} \prod_{t=1}^T \qtzgxsy \log \prod_{t=1}^T \ptxgz \nonumber\\
&=& \sum_{t=1}^T \int_{\bb{z}_{1:T}} \prod_{t=1}^T \qtzgxsy \log \ptxgz \nonumber\\
&=& \sum_{t=1}^T \E{\qtzgxsy}{\log \ptxgz}.
\end{eqnarray}
For the second term,
\begin{align} \label{eq:rslds_elbo_sup_2}
&\E{q\left(\bb{z}_{1:T} | \bb{x}_{1:T}, y_{1:T}\right)}{\log \frac{p\left(\bb{z}_{1:T}, y_{1:T}\right)}{q\left(\bb{z}_{1:T} | \bb{x}_{1:T}, y_{1:T}\right)}} = \nonumber\\[1ex]
&\int_{\bb{z}_{1:T}} \prod_{t=1}^T \qtzgxsy \log \frac{\poz \poy \prod_{t=2}^T \ptzgzmoy \ptygymozmo}{\prod_{t=1}^T \qtzgxsy} \nonumber\\[1ex]
=& \int_{\bb{z}_{1:T}} \prod_{t=1}^T \qtzgxsy \log \frac{\poz \poy}{q\left(\bb{z}_1 | \bb{x}_{\mathcal{T}_1}, y_1\right)} \nonumber\\[1ex]
&+ \sum_{t=2}^T \int_{\bb{z}_{1:T}} \prod_{t=1}^T \qtzgxsy \log \frac{\ptzgzmoy \ptygymozmo}{\qtzgxsy} \nonumber\\[1ex]
=& \int_{\bb{z}_1} q\left(\bb{z}_1 | \bb{x}_{\mathcal{T}_1}, y_1\right) \log \frac{\poz \poy}{q\left(\bb{z}_1 | \bb{x}_{\mathcal{T}_1}, y_1\right)} \nonumber\\[1ex]
&+ \sum_{t=2}^T \int_{\bb{z}_{t-1}} \int_{\bb{z}_{t}} \qtmozgxsy \qtzgxsy \log \frac{\ptzgzmoy \ptygymozmo}{\qtzgxsy} \nonumber\\[1ex]
=& \log \poy -\KL\left[q\left(\bb{z}_1 | \bb{x}_{\mathcal{T}_1}, y_1\right) || \poz \right] - \sum_{t=2}^T \E{\qtmozgxsy}{\KL\left[ \qtzgxsy || \ptzgzmoy \right]} \nonumber\\[1ex]
&+ \sum_{t=2}^T \E{\qtmozgxsy}{\log \ptygymozmo}.
\end{align}
Now we combine Eqs. \ref{eq:rslds_elbo_sup_1} and \ref{eq:rslds_elbo_sup_2} to get the full variational lower bound:
\begin{align} \label{eq:rslds_elbo_sup}
\log p\left(\bb{x}_{1:T}, y_{1:T}\right) \geq & \sum_{t=1}^T \E{\qtzgxsy}{\log \ptxgz} \nonumber\\
&- \KL\left[q\left(\bb{z}_1 | \bb{x}_{\mathcal{T}_1}, y_1\right) || \poz \right] + \log \poy\nonumber\\
&- \sum_{t=2}^T \E{\qtmozgxsy}{\KL\left[ \qtzgxsy || \ptzgzmoy \right]} \nonumber\\
&+ \sum_{t=2}^T \E{\qtmozgxsy}{\log \ptygymozmo} \\
\approx & \sum_{t=1}^T {\log p\left(\bb{x}_t | \Tilde{\bb{z}}_t\right)} \nonumber \\
&- \KL\left[q\left(\bb{z}_1 | \bb{x}_{\mathcal{T}_1}, y_1\right) || \poz \right]
+ \log \poy \nonumber \\
&- \sum_{t=2}^T {\KL\left[ \qtzgxsy || p\left(\bb{z}_t | \Tilde{\bb{z}}_{t-1}, y_t\right) \right]} \nonumber\\
&+ \sum_{t=2}^T \log p\left(y_t | y_{t-1}, \Tilde{\bb{z}}_{t-1}\right).
\end{align}

\subsection{All labels unobserved}
Next, we derive the variational lower bound for the more complicated case where we assume all $y_t$ are \textit{un}observed.
\begin{align}
\log p\left(\bb{x}_{1:T}\right) \geq & \int_{\bb{z}_{1:T}} \int_{y_{1:T}} q\left(\bb{z}_{1:T}, y_{1:T} | \bb{x}_{1:T}\right) \log \frac{p\left(\bb{x}_{1:T} | \bb{z}_{1:T}\right)p\left(\bb{z}_{1:T}, y_{1:T}\right)}{q\left(\bb{z}_{1:T}, y_{1:T} | \bb{x}_{1:T}\right)} \nonumber\\
=& \int_{\bb{z}_{1:T}} \int_{y_{1:T}} q\left(\bb{z}_{1:T}, y_{1:T} | \bb{x}_{1:T}\right) \log p\left(\bb{x}_{1:T} | \bb{z}_{1:T}\right) \nonumber\\
&+ \int_{\bb{z}_{1:T}} \int_{y_{1:T}} q\left(\bb{z}_{1:T}, y_{1:T} | \bb{x}_{1:T}\right) \log \frac{p\left(\bb{z}_{1:T}, y_{1:T}\right)}{q\left(\bb{z}_{1:T}, y_{1:T} | \bb{x}_{1:T}\right)}\nonumber
\end{align}

Again, recall that both $q\left(\bb{z}_{1:T} | \bb{x}_{1:T}, y_{1:T}\right)$ and $p\left(\bb{x}_{1:T} | \bb{z}_{1:T}\right)$ factorize across time. We now expand the first term:
\begin{align} \label{eq:rslds_elbo_unsup_1}
&\int_{\bb{z}_{1:T}} \int_{y_{1:T}} q\left(\bb{z}_{1:T}, y_{1:T} | \bb{x}_{1:T}\right) \log p\left(\bb{x}_{1:T} | \bb{z}_{1:T}\right) =\nonumber\\[1ex]
& \int_{\bb{z}_{1:T}} \int_{y_{1:T}} \prod_{t=1}^T \qtygxs \qtzgxsy \log \prod_{t=1}^T \ptxgz \nonumber\\[1ex]
=& \sum_{t=1}^T \int_{\bb{z}_{1:T}} \int_{y_{1:T}} \prod_{t=1}^T \qtygxs \qtzgxsy \log \ptxgz \nonumber\\[1ex]
=& \sum_{t=1}^T \int_{\bb{z}_{t}} \int_{y_t} \qtygxs \qtzgxsy \log \ptxgz \nonumber\\[1ex]
=& \sum_{t=1}^T \E{\qtygxs \qtzgxsy}{\log \ptxgz}.
\end{align}

For the second term,
\begin{align} \label{eq:rslds_elbo_unsup_2}
&\int_{\bb{z}_{1:T}} \int_{y_{1:T}} q\left(\bb{z}_{1:T}, y_{1:T} | \bb{x}_{1:T}\right) \log \frac{p\left(\bb{z}_{1:T}, y_{1:T}\right)}{q\left(\bb{z}_{1:T}, y_{1:T} | \bb{x}_{1:T}\right)}\nonumber\\[1ex]
=& \int_{\bb{z}_{1:T}} \int_{y_{1:T}} \prod_{t=1}^T \qtygxs \qtzgxsy \log \frac{\poz \poy \prod_{t=2}^T \ptzgzmoy \ptygymozmo}{\prod_{t=1}^T \qtygxs \qtzgxsy}\nonumber\\[1ex]
=& \int_{\bb{z}_{1:T}} \int_{y_{1:T}} \prod_{t=1}^T \qtygxs \qtzgxsy \log \frac{\poz \poy}{q\left(\bb{z}_1 | \bb{x}_{\mathcal{T}_1}, y_1\right)q\left(y_1|\bb{x}_{\mathcal{T}_1}\right)}\nonumber\\[1ex]
&+ \int_{\bb{z}_{1:T}} \int_{y_{1:T}} \prod_{t=1}^T \qtygxs \qtzgxsy \left[ \sum_{t=2}^T \log \frac{\ptzgzmoy \ptygymozmo}{\qtygxs \qtzgxsy}\right]\nonumber\\[1ex]
=& \int_{\bb{z}_{1}} \int_{y_{1}} q\left(\bb{z}_1 | \bb{x}_{\mathcal{T}_1}, y_1\right)q\left(y_1|\bb{x}_{\mathcal{T}_1}\right) \log \frac{\poz \poy}{q\left(\bb{z}_1 | \bb{x}_{\mathcal{T}_1}, y_1\right)q\left(y_1|\bb{x}_{\mathcal{T}_1}\right)}\nonumber\\[1ex]
&+ \sum_{t=2}^T \int_{\bb{z}_{t-1}} \int_{\bb{z}_{t}} \int_{y_{t-1}} \int_{y_{t}} \qtygxs \qtmoygxs \qtzgxsy \qtmozgxsy \log \frac{\ptzgzmoy}{\qtzgxsy}\nonumber\\[1ex]
&+ \sum_{t=2}^T \int_{\bb{z}_{t-1}} \int_{\bb{z}_{t}} \int_{y_{t-1}} \int_{y_{t}} \qtygxs \qtmoygxs \qtzgxsy \qtmozgxsy \log \frac{\ptygymozmo}{\qtygxs}.
\end{align}

The first term from Eq. \ref{eq:rslds_elbo_unsup_2} becomes
\begin{align} \label{eq:rslds_elbo_unsup_2a}
&\int_{\bb{z}_{1}} \int_{y_{1}} q\left(\bb{z}_1 | \bb{x}_{\mathcal{T}_1}, y_1\right)q\left(y_1|\bb{x}_{\mathcal{T}_1}\right) \log \frac{\poz \poy}{q\left(\bb{z}_1 | \bb{x}_{\mathcal{T}_1}, y_1\right)q\left(y_1|\bb{x}_{\mathcal{T}_1}\right)}\nonumber\\[1ex]
=& \int_{y_{1}} q\left(y_1|\bb{x}_{\mathcal{T}_1}\right) \int_{\bb{z}_{1}} q\left(\bb{z}_1 | \bb{x}_{\mathcal{T}_1}, y_1\right) \log \frac{\poz}{q\left(\bb{z}_1 | \bb{x}_{\mathcal{T}_1}, y_1\right)} + \int_{y_{1}} q\left(y_1|\bb{x}_{\mathcal{T}_1}\right) \log \frac{\poy}{q\left(y_1 | \bb{x}_{\mathcal{T}_1}\right)}   \int_{\bb{z}_{1}} q\left(\bb{z}_1 | \bb{x}_{\mathcal{T}_1}, y_1\right) \nonumber\\[1ex]
=& -\E{q\left(y_1 | \bb{x}_{\mathcal{T}_1}\right)}{\KL\left[q\left(\bb{z}_1 | \bb{x}_{\mathcal{T}_1}, y_1\right) || \poz \right]} - \KL \left[ q\left(y_1 | \bb{x}_{\mathcal{T}_1}\right) || \poy \right].
\end{align}

The second term from Eq. \ref{eq:rslds_elbo_unsup_2} becomes
\begin{align} \label{eq:rslds_elbo_unsup_2b}
&\sum_{t=2}^T \int_{\bb{z}_{t-1}} \int_{\bb{z}_{t}} \int_{y_{t-1}} \int_{y_{t}} \qtygxs \qtmoygxs \qtzgxsy \qtmozgxsy \log \frac{\ptzgzmoy}{\qtzgxsy}\nonumber\\[1ex]
=& - \sum_{t=2}^T \E{\qtygxs \qtmoygxs \qtmozgxsy}{\KL\left[ \qtzgxsy || \ptzgzmoy \right]}
\end{align}

and the third term from Eq. \ref{eq:rslds_elbo_unsup_2} becomes
\begin{align} \label{eq:rslds_elbo_unsup_2c}
&\sum_{t=2}^T \int_{\bb{z}_{t-1}} \int_{y_{t-1}} \int_{y_{t}} \qtygxs \qtmoygxs \qtmozgxsy \log \frac{\ptygymozmo}{\qtygxs}\nonumber\\[1ex]
=& - \sum_{t=2}^T \E{\qtmoygxs \qtmozgxsy}{\KL\left[ \qtygxs || \ptygymozmo \right]}.
\end{align}

Now we combine Eqs. \ref{eq:rslds_elbo_unsup_1}, \ref{eq:rslds_elbo_unsup_2a}, \ref{eq:rslds_elbo_unsup_2b}, and \ref{eq:rslds_elbo_unsup_2c} to get the full variational lower bound:

\begin{align} \label{eq:rslds_elbo_unsup_marg}
\log p\left(\bb{x}_{1:T}\right) \geq & \sum_{t=1}^T \E{\qtygxs \qtzgxsy}{\log \ptxgz}\nonumber\\[1ex]
&-\E{q\left(y_1 | \bb{x}_{\mathcal{T}_1}\right)}{\KL\left[q\left(\bb{z}_1 | \bb{x}_{\mathcal{T}_1}, y_1\right) || \poz \right]} - \KL \left[ q\left(y_1 | \bb{x}_{\mathcal{T}_1}\right) || \poy \right]\nonumber\\[1ex]
&- \sum_{t=2}^T \E{\qtygxs \qtmoygxs \qtmozgxsy}{\KL\left[ \qtzgxsy || \ptzgzmoy \right]} \nonumber\\[1ex]
&- \sum_{t=2}^T \E{\qtmoygxs \qtmozgxsy}{\KL\left[ \qtygxs || \ptygymozmo \right]} \nonumber\\
= & \sum_{t=1}^T \sum_{k=1}^K  \Big( q\left(y_t = k | \bb{x}_{\mathcal{T}_t}\right) \mathbb{E}_{q\left(\bb{z}_t|\bb{x}_{\mathcal{T}_t}, y_t=k\right)} [{\log p\left(\bb{x}_t | {\bb{z}}_t\right)}] \Big) \nonumber \\
&- \sum_{k=1}^{K} \Big( q\left(y_1 =k | \bb{x}_{\mathcal{T}_1}\right) {\KL\left[q\left(\bb{z}_1 | \bb{x}_{\mathcal{T}_1}, y_1 =k\right) || \poz \right]}\Big) - \KL \left[ q\left(y_1 | \bb{x}_{\mathcal{T}_1}\right) || \poy \right]\nonumber\\
&- \sum_{t=2}^T \sum_{k=1}^{K} \sum_{k'=1}^{K} \Big( q\left(y_t =k | \bb{x}_{\mathcal{T}_t}\right) q\left(y_{t-1} =k' | \bb{x}_{\mathcal{T}_{t-1}}\right)\nonumber\\ & \quad \quad \mathbb{E}_{q\left(\bb{z}_{t-1}|\bb{x}_{\mathcal{T}_{t-1}}, y_{t-1}=k'\right)}[{\KL\left[ {q\left(\bb{z}_t | \bb{x}_{\mathcal{T}_t}, y_t =k\right)} || {p\left(\bb{z}_t | \bb{z}_{t-1},y_t = k\right)} \right]} ]\Big) \nonumber\\[1ex]
&- \sum_{t=2}^T \sum_{k=1}^{K} \Big( q\left(y_{t-1} =k | \bb{x}_{\mathcal{T}_{t-1}}\right)\mathbb{E}_{q\left(\bb{z}_{t-1}|\bb{x}_{\mathcal{T}_{t-1}}, y_{t-1}=k\right)}[{\KL\left[ \qtygxs || {p\left(y_t | y_{t-1} =k, \bb{z}_{t-1}\right)} \right]}] \Big) \\
\approx & \sum_{t=1}^T \sum_{k=1}^K  \Big( q\left(y_t = k | \bb{x}_{\mathcal{T}_t}\right) {\log p\left(\bb{x}_t | {\tilde{\bb{z}}}_t^k\right)} \Big)  \nonumber \\
&- \sum_{k=1}^{K} \Big( q\left(y_1 =k | \bb{x}_{\mathcal{T}_t}\right) {\KL\left[q\left(\bb{z}_1 | \bb{x}_{\mathcal{T}_1}, y_1 =k\right) || \poz \right]}\Big) - \KL \left[ q\left(y_1 | \bb{x}_{\mathcal{T}_1}\right) || \poy \right]\nonumber\\
&- \sum_{t=2}^T \sum_{k=1}^{K} \sum_{k'=1}^K \Big( q\left(y_t =k | \bb{x}_{\mathcal{T}_t}\right) q\left(y_{t-1} =k' | \bb{x}_{\mathcal{T}_{t-1}}\right)\nonumber\\
& \quad \quad{\KL\left[ {q\left(\bb{z}_t | \bb{x}_{\mathcal{T}_t}, y_t =k\right)} || {p\left(\bb{z}_t | \tilde{\bb{z}}_{t-1}^{k'},y_t = k\right)} \right]} \Big) \nonumber \\[1ex]
&- \sum_{t=2}^T \sum_{k=1}^{K} \Big( q\left(y_{t-1} =k | \bb{x}_{\mathcal{T}_{t-1}}\right){\KL\left[ \qtygxs || {p\left(y_t | y_{t-1} =k, \tilde{\bb{z}}_{t-1}^k\right)} \right]} \Big) \label{eq:rslds_elbo_unsup_marg_f2}
\end{align}

where $$\Tilde{\bb{z}}_t^k \sim q\left(\bb{z}_t | \bb{x}_{\mathcal{T}_t}, y_t =k\right)\ \textrm{and} 
 \ \Tilde{\bb{z}}_{t-1}^{k} \sim q\left(\bb{z}_{t-1} | \bb{x}_{\mathcal{T}_{t-1}}, y_{t-1} =k\right).$$

\subsection{Mix of observed and unobserved labels}

The loss function that incorporates both observed and unobserved discrete labels can be computed in an elegant and straightforward manner. For notational simplicity we define a new distribution over the discrete variables, which is equal to the approximate posterior for unlabeled time points, and equal to a one-hot vector representing the true discrete state for labeled time points. Let $\mathcal{T}_{\mathcal{L}}$ be the set of all labeled time points and $\mathcal{T}_{\mathcal{U}}$ be the set of all unlabeled time points, such that $\mathcal{T}_{\mathcal{L}} \cup \mathcal{T}_{\mathcal{U}} = \{1, \ldots, T\}$ and $\mathcal{T}_{\mathcal{L}} \cap \mathcal{T}_{\mathcal{U}} = \varnothing$. Then we define
\begin{eqnarray*}
r\left(y_t|\bb{x}_{\mathcal{T}_t}\right) =
\begin{cases}
    e_{y_t} & \text{if } t \in \mathcal{T}_{\mathcal{L}}\\
    q\left(y_t| \bb{x}_{\mathcal{T}_t}\right) & \text{if } t \in \mathcal{T}_{\mathcal{U}}\\
\end{cases}
\end{eqnarray*}
where $e_{y_t}$ denotes a vector with a 1 in the $y_t^{\text{th}}$ coordinate and 0 elsewhere. We can then compute Eq.~\ref{eq:rslds_elbo_unsup_marg_f2} using this mixture of distributions due to the fact that each of the individual terms in the unsupervised ELBO (Eq.~\ref{eq:rslds_elbo_unsup_marg_f}) reduces to the corresponding term in the supervised ELBO (Eq.~\ref{eq:rslds_elbo_sup_f}) when using one-hot vectors. In addition, we compute the classification loss on the approximate posterior $q\left(y_t | \bb{x}_{\mathcal{T}_t}\right)$ for all observed discrete labels:
\begin{align} \label{eq:rslds_elbo_semisup_marg}
\mathcal{L}_{ss} = & \sum_{t=1}^T \sum_{k=1}^K  \Big( r\left(y_t = k | \bb{x}_{\mathcal{T}_t}\right) {\log p\left(\bb{x}_t | {\tilde{\bb{z}}}_t^k \right)} \Big)  \nonumber \\
&- \sum_{k=1}^{K} \Big( r\left(y_1 =k | \bb{x}_{\mathcal{T}_t}\right) {\KL\left[q\left(\bb{z}_1 | \bb{x}_{\mathcal{T}_1}, y_1 =k\right) || \poz \right]}\Big) - \KL \left[ r\left(y_1 | \bb{x}_{\mathcal{T}_1}\right) || \poy \right]\nonumber\\
&- \sum_{t=2}^T \sum_{k=1}^{K} \sum_{k'=1}^K \Big( r\left(y_t =k | \bb{x}_{\mathcal{T}_t}\right) r\left(y_{t-1} =k' | \bb{x}_{\mathcal{T}_{t-1}}\right)\nonumber\\
& \quad \quad{\KL\left[ {q\left(\bb{z}_t | \bb{x}_{\mathcal{T}_t}, y_t =k\right)} || {p\left(\bb{z}_t | \tilde{\bb{z}}_{t-1}^{k'},y_t = k\right)} \right]} \Big) \nonumber \\[1ex]
&- \sum_{t=2}^T \sum_{k=1}^{K} \Big( r\left(y_{t-1} =k | \bb{x}_{\mathcal{T}_{t-1}}\right){\KL\left[ r\left(y_t | \bb{x}_{\mathcal{T}_t}\right) || {p\left(y_t | y_{t-1} =k, \tilde{\bb{z}}_{t-1}^k\right)} \right]} \Big) \nonumber \\
&+ \sum_{t \in \mathcal{T}_{\mathcal{L}}} \alpha \log q\left(y_t | \bb{x}_{\mathcal{T}_t}\right).
\end{align}

\newpage
\clearpage

\section{The International Brain Laboratory consortium members} \label{app:ibl}

Larry Abbot$^{1}$,
Luigi Acerbi$^{2}$, 
Valeria Aguillon-Rodriguez$^{3}$,
Mandana Ahmadi$^{4}$, 
Jaweria Amjad$^{4}$, 
Dora Angelaki$^{5}$, 
Jaime Arlandis$^{6}$, 
Zoe C Ashwood$^{7}$, 
Kush Banga$^{8}$, 
Hailey Barrell$^{9}$
% ...

\small{$^{1}$Zuckerman Institute, Columbia University, New York, NY, USA}\\
\small{$^{2}$Department of Basic Neuroscience, University of Geneva, Geneva, Switzerland}\\
\small{$^{3}$Cold Spring Harbor Laboratory, Cold Spring Harbor, NY, USA}\\
\small{$^{4}$Gatsby Computational Neuroscience Unit, University College London, London, UK}\\
\small{$^{5}$Center for Neural Science, New York University, New York, NY, USA}\\
\small{$^{6}$Champalimaud Center for the Unknown, Lisboa, Portugal}\\
\small{$^{7}$Princeton Neuroscience Institute, Princeton University, Princeton, NJ, USA}\\
\small{$^{8}$Institute of Neurology, University College London, London, UK}\\
\small{$^{9}$Department of Biological Structure, University of Washington, Seattle, WA, USA}\\
\small{$^{10}$Department of Applied Physics, Stanford University, Stanford, CA, USA}\\
\small{$^{11}$Max Planck Institute for Biological Cybernetics, Tübingen, Germany}\\
\small{$^{12}$Sainsbury-Wellcome Centre, University College London, London, UK}\\
\small{$^{13}$Institute of Opthalmology, University College London, London, United Kingdom}\\
\small{$^{14}$Department of Neurobiology, University of California, Los Angeles, CA, USA}\\
\small{$^{15}$Department of  Molecular and Cell Biology, University of California, Berkeley, CA, USA}\\
\small{$^{16}$Département D’études Cognitives, École Normale Supérieure, Paris, France}\\
\small{$^{17}$Department of Brain and Cognitive Sciences, Massachusetts Institute of Technology, Cambridge, MA, USA}\\
\small{$^{18}$Wolfson Institute of Biomedical Research, University College London, London, United Kingdom}\\
\small{$^{19}$Center for Computational Neuroscience, University of Washington, Seattle, WA, USA}\\
\small{$^{20}$Department of Physiology, University of Yamanashi, Kofu, Yamanashi, Japan}\\
\small{$^{21}$The Allen Institute for Neural Dynamics, Seattle, Washington, USA}\\

\newpage
\clearpage

\end{document}